\pdfoutput=1

\documentclass[11pt]{article}

\usepackage[final]{acl}

\usepackage{times}
\usepackage{latexsym}

\usepackage[T1]{fontenc}

\usepackage[utf8]{inputenc}

\usepackage{microtype}

\usepackage{inconsolata}

\usepackage{graphicx}

\usepackage{multirow}
\usepackage{booktabs}
\usepackage{enumitem}

\definecolor{lightgray}{gray}{0.9}
\definecolor{nmgray}{RGB}{229,229,229}

\usepackage{xspace}
\usepackage{makecell}
\usepackage{amsmath}
\usepackage{cleveref}
\usepackage{framed}
\usepackage{pifont} 
\usepackage{hyperref} 

\usepackage{listings} 
\usepackage{tcolorbox}
\lstdefinestyle{prompt}{
    frame=single,
    basicstyle=\footnotesize\ttfamily,
    columns=fullflexible,
    breaklines=true,
    extendedchars=true,
    escapechar=@,
    literate={á}{{\'a}}1 {ã}{{\~a}}1 {é}{{\'e}}1 {£}{{\pounds}}1 {–}{{-}}1 {’}{{'}}1,
}

\newcommand{\etc}{\emph{etc}\xspace}
\newcommand{\eg}{\emph{e.g.}\xspace}
\newcommand{\ie}{\emph{i.e.}\xspace}

\definecolor{firstBest}{rgb}{0.86, 1, 0.86} 

\definecolor{kellygreen}{rgb}{0.3, 0.73, 0.09}
\definecolor{alizarin}{rgb}{0.82, 0.1, 0.26}
\newcommand{\cmark}{{\color{kellygreen} \ding{51}}}
\newcommand{\xmark}{{\color{alizarin} \ding{55}}}

\def\locating{locating}
\def\Locating{Locating}

\def\benchName{LongDocURL}
\def\BenchName{LongDoc\textbf{URL}}
\title{\benchName: a Comprehensive Multimodal Long Document Benchmark Integrating \emph{Understanding}, \emph{Reasoning}, and \emph{\Locating}}
  
\author{
    Chao Deng$^{1,2,3}$\thanks{~~~Equal contribution.}\thanks{~~~Work done during an internship at Alibaba Group.}, Jiale Yuan$^{4}$\footnotemark[1], Pi Bu$^{4}$, Peijie Wang$^{1,2}$, Zhong-Zhi Li$^{1,2}$, Jian Xu$^{1,2}$, \\
    \textbf{Xiao-Hui Li$^{1,2}$, Yuan Gao$^{4}$, Jun Song$^{4}$\footnotemark[3], Bo Zheng$^{4}$, Cheng-Lin Liu$^{1,2,3}$\thanks{~~~Corresponding authors.}} \\
    MAIS, Institute of Automation, Chinese Academy of Sciences$^{1}$ \\
    School of Artificial Intelligence, University of Chinese Academy of Sciences$^{2}$ \\
    Zhongguancun Academy, Beijing$^{3}$ \qquad
    Taobao \& Tmall Group of Alibaba$^{4}$ \\
    dengchao2023@ia.ac.cn, liucl@nlpr.ia.ac.cn  \\
    \{yuanjiale.yjl, bupi.wj, jsong.sj\}@alibaba-inc.com
}

\begin{document}
\maketitle

\begin{abstract}
\label{sec:abstract}
Large vision language models (LVLMs) have improved the document understanding capabilities remarkably, enabling the handling of complex document elements, longer contexts, and a wider range of tasks. However, existing document understanding benchmarks have been limited to handling only a small number of pages and fail to provide a comprehensive analysis of layout elements locating. In this paper, we first define three primary task categories: \textbf{Long} \textbf{Doc}ument \textbf{U}nderstanding, numerical \textbf{R}easoning, and cross-element \textbf{L}ocating, and then propose a comprehensive benchmark—\textbf{\benchName}—integrating above three primary tasks and comprising 20 sub-tasks categorized based on different primary tasks and answer evidences. Furthermore, we develop a semi-automated construction pipeline and collect 2,325 high-quality question-answering pairs, covering more than 33,000 pages of documents, significantly outperforming existing benchmarks. Subsequently, we conduct comprehensive evaluation experiments on both open-source and closed-source models across 26 different configurations, revealing critical performance gaps in this field. The code and data: \url{https://github.com/dengc2023/LongDocURL}.

\end{abstract}   
\section{Introduction}
\label{sec:intro}

The research of document understanding has been advanced remarkably in the last decade. However, past works mostly rely on smaller, specialized models, necessitating the design of independent models for each specific task (\eg, document structure parsing). This strategy not only increases the labor in model development but also limits the applicability of the models. Recently, this field has undergone transformation with the rise of large language models (LLMs) and large vision-language models (LVLMs), such as the open-source InternLM-XC2-4KHD~\cite{dong2024internlm} and TextMonkey~\cite{liu2024textmonkey}. These models are showcasing their capabilities in handling complex document elements like charts and images, managing longer contexts up to 128k or more, and tackling diverse tasks in addition to OCR task, such as table question answering and layout understanding.

\begin{figure}
    \centering
    \includegraphics[width=0.5\textwidth]{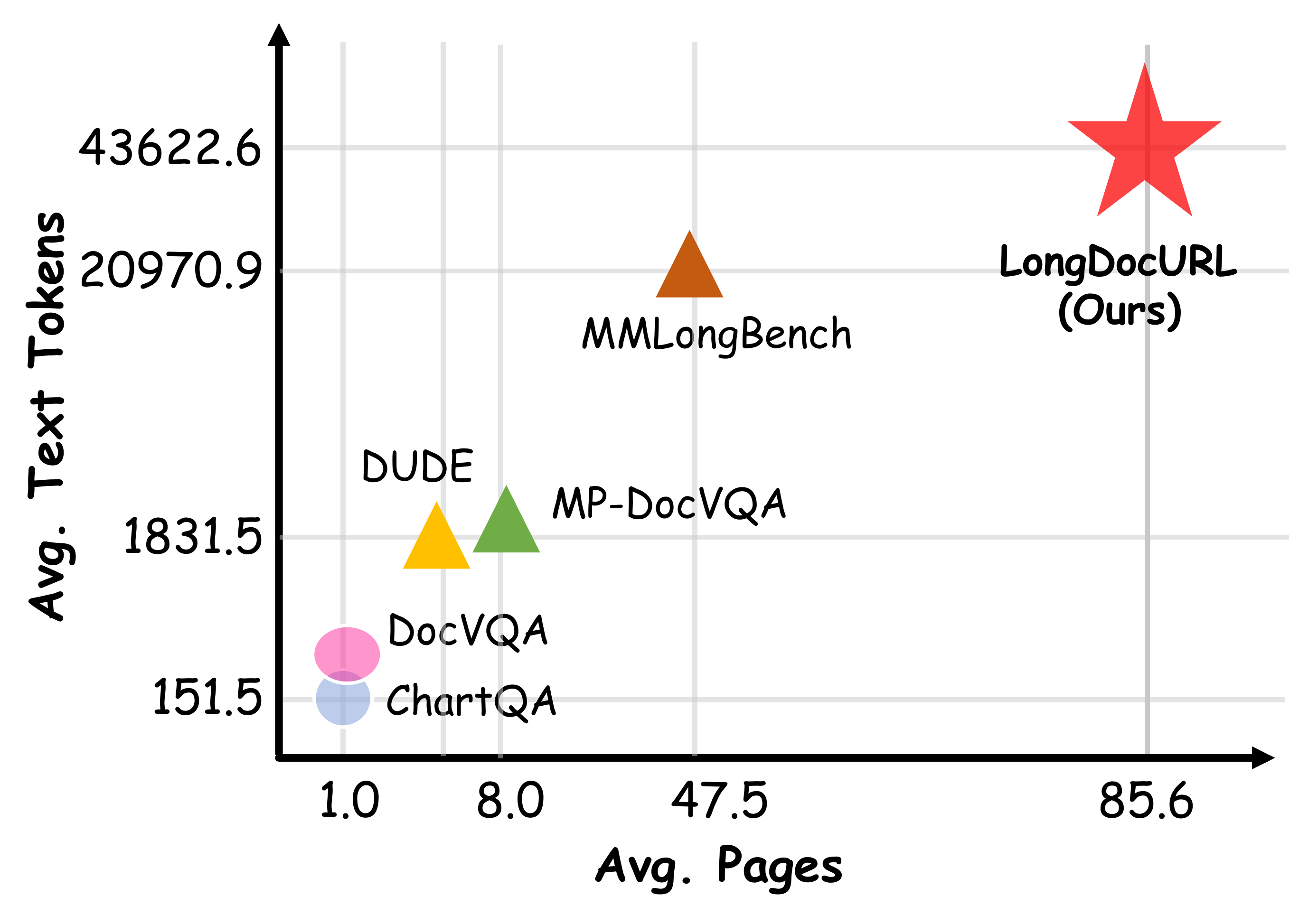} 
    \vspace{-0.8cm}
    \caption{Comparison with other datasets in average pages and text tokens per document.}
    \vspace{-0.8cm}
    \label{fig:comparison}
\end{figure}

\begin{figure*}[ht]
    \centering
    \includegraphics[width=1.0\textwidth]{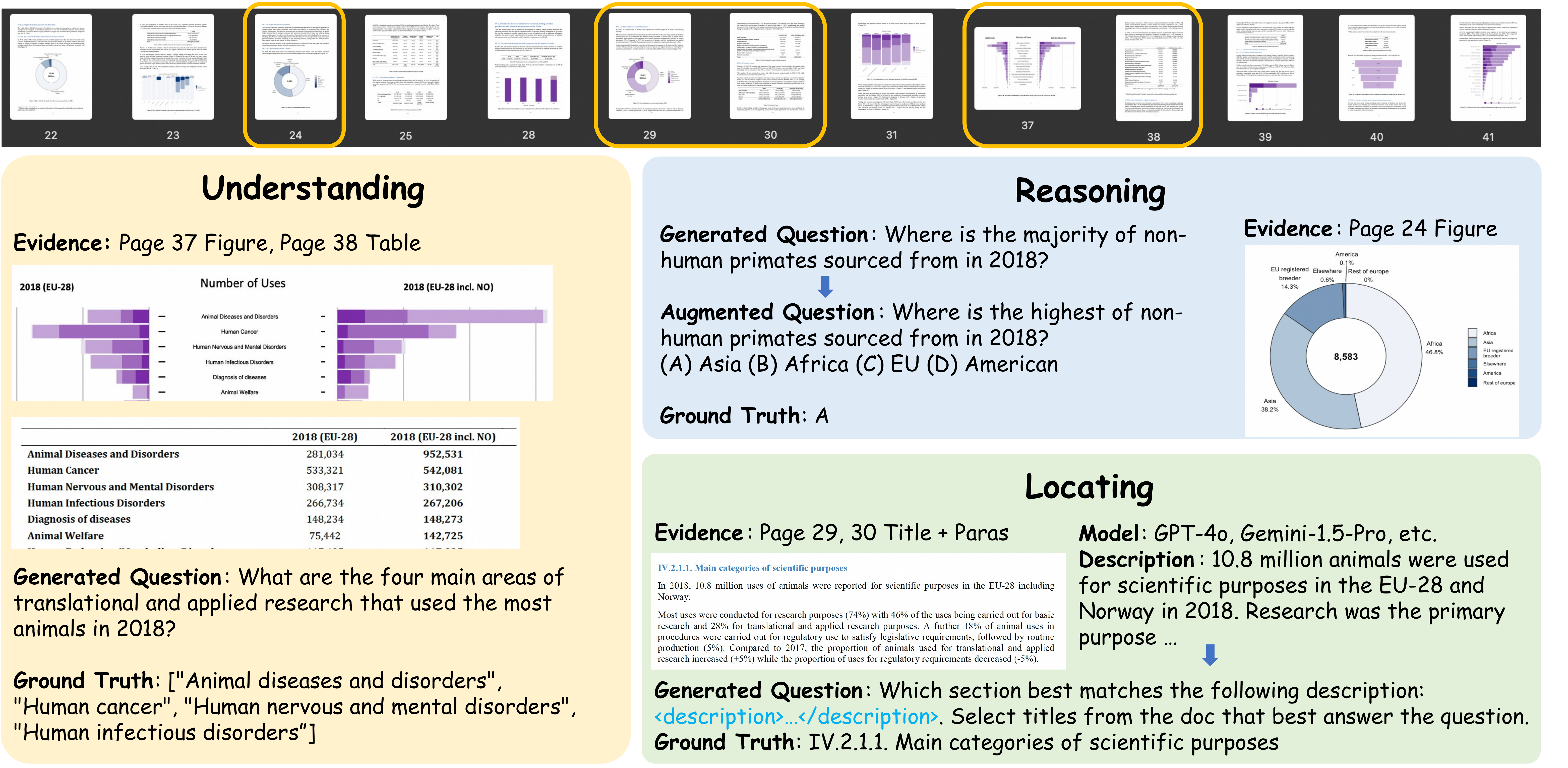} 
    \vspace{-0.7cm}
    \caption{\benchName~comprises 20 sub-tasks focusing on three task categories: \textbf{U}nderstanding, numerical \textbf{R}easoning, and cross-element \textbf{L}ocating. (Top) Thumbnail of a document example. \textcolor{orange}{Orange} boxes indicate answer evidence pages. (Bottom) Data examples generated from the document and screenshots of relevant part of answer evidence pages.}
    \vspace{-0.5cm}
    \label{fig:overview}
\end{figure*}

Despite the advances in model capabilities, the evaluation of complex document tasks is somewhat lacking. Take DocVQA~\cite{DocVQA} as an example; it is one of the standard benchmarks for document understanding, but it can only assess single-page documents, and many models can easily exceed an accuracy of 95\% (\eg, Qwen2-VL~\cite{Qwen2-VL}). The benchmarks detailed in \Cref{tab:comparison_among_benchmarks} exhibit limitations in adequately addressing complex elements, longer contexts, and diverse tasks: \textbf{1) Complex elements:} Most benchmarks fail to cover all elements such as paragraphs, titles, tables, and figures, focusing instead on only some of the contents. Additionally, discussions about the interrelations among different elements are scarce. \textbf{2) Longer contexts:} Current benchmarks for multi-page document question answering, such as MP-DocVQA~\cite{MP-DocVQA} and DUDE~\cite{DUDE}, do not assess documents exceeding 20 pages. While MMLongBench-Doc~\cite{MMLongBench-Doc} collects longer documents, it offers only approximately 1k effective samples, with only about 30\% of questions involving cross-page information. \textbf{3) Diverse tasks:} Existing work focuses more on OCR or easy question-answering tasks, neglecting the exploration of capabilities in other areas such as cross-element \locating~task (as shown in \Cref{fig:overview}c). The above findings indicate that existing benchmarks lag behind the advances of models, which could hinder the development of document understanding.

\begin{table*}[htbp]
    \centering
    \resizebox{0.95\textwidth}{!}{
    \begin{tabular}{l|cccc|cc|ccc}
        \toprule
        
        \multirow{2}{*}{\textbf{Benchmarks}} & \multicolumn{4}{c|}{\textbf{Data Size}} & \multicolumn{2}{c|}{\textbf{Answer Evidence}} & \multicolumn{3}{c}{\textbf{Task Type}} \\
        
         & \#Docs & \#Avg. Pages & \#Avg. Tokens & \#QA & Multi-page(\%) & Cross-element(\%) & U & R & L \\
        
        \midrule
        
        \multicolumn{2}{l}{\textcolor{gray}{\textit{single-page}}} \\
        DocVQA~\cite{DocVQA} & - & 1.0 & 151.5 & - & \xmark & not specified & \cmark & \xmark & \xmark \\
        ChartQA~\cite{ChartQA} & - & 1.0 &  236.9 & - & \xmark & not specified & \cmark & \xmark & \xmark\\
        
        \multicolumn{2}{l}{\textcolor{gray}{\textit{multi-page(<=20)}}} \\
        MP-DocVQA~\cite{MP-DocVQA} & - & 8.3 & 2,026.6 & - & \xmark & not specified & \cmark & \xmark & \xmark\\
        DUDE~\cite{DUDE} & - & 5.7 & 1,831.5 & - & \cmark(2.1\%) & not specified & \cmark & \cmark & \xmark\\
        
        \multicolumn{2}{l}{\textcolor{gray}{\textit{multi-page(>20)}}} \\
        MMLongBench-Doc~\cite{MMLongBench-Doc} & 135 & 47.5 & 21,214.1 & 1,082 & \cmark(33.0\%) & \cmark(22.6\%) & \cmark & \cmark & \xmark \\
        M-Longdoc~\cite{M-Longdoc-arxiv} & 180 & 210.8 & - & 851 & - & - & \cmark & \cmark & \xmark \\
        
        \midrule
        
        \benchName~(Ours) & 396 & 85.6 & 43,622.6 & 2,325 & \cmark(52.9\%) & \cmark(37.1\%) & \cmark & \cmark & \cmark\\
        \bottomrule
    \end{tabular}}
    \caption{Comparison between \BenchName~and previous document understanding datasets. Task types: (\textbf{U})nderstanding, (\textbf{R})easoning, and (\textbf{L})ocating.}
    \label{tab:comparison_among_benchmarks}
    \vspace{-5mm}
\end{table*}

In this paper, we present a comprehensive document benchmark including three task categories: 1) \textbf{Understanding}: extracting information from documents by identifying keywords, parsing the structure of tables, \etc. Answers are found directly in the document. 2) Numerical \textbf{Reasoning}: processing numerical information through counting, calculating, comparing, and summarizing, requiring both extracted information and reasoning for concluding. 3) Cross-Element \textbf{Locating}: As mentioned earlier, discussions about the interrelations among different types of elements are scarce. It is often necessary to establish a task that evaluates models' ability to analyze relations among different types of elements. For instance, in Para-Title \Locating~task, as shown in \Cref{fig:overview}c, models must summarize relevant sections to identify parts that match a given abstract and then determine the relation between the paragraph and its section titles. This task requires switching element types (\ie, paragraphs to titles) during the answering process.

Our benchmark, named \textbf{\benchName}, comprises 20 sub-tasks according to different primary tasks and answer evidence. More details are presented in \Cref{sec:benchmark}. To efficiently assemble the evaluation dataset for \benchName, we design a semi-automated pipeline comprising four modules. Specifically, a Extract \& Filter module identifies documents of suitable length with rich layouts from diverse document sources. A QA Generation module utilizes a multi-step iterative querying process with advanced models (\eg, GPT-4o) to generate QA pairs with evidence sources. Finally, the Automated Verification and Human Verification modules ensure the quality of the generated content. Through this semi-automated pipeline, we ultimately produce 2,325 QA pairs, covering more than 33,000 pages of documents. Thereafter, we conduct comprehensive evaluation experiments with 26 different configurations (varying the model and input format). These evaluation results indicate that the highest-performing closed-source model, GPT-4o, scored 64.5, leading all models, while the best score for open-source models is only 30.6. This result reveals a potential gap in document understanding and shows the need for further improvement. Our contributions are as follows:
\begin{itemize}[leftmargin=*]
    \item We introduce three primary tasks of long document and propose a comprehensive benchmark comprising 20 sub-tasks categorized based on different primary tasks and answer evidences to support more fine-grained evaluation.
    \item We develop a cost-efficient semi-automated construction pipeline and generate 2,325 high-quality QA pairs, covering more than 33,000 pages of documents, which significantly outperforms existing benchmarks.
    \item We conduct comprehensive evaluation experiments of both open-source and closed-source models under 26 different configurations.
\end{itemize}
\section{Related Work}
\label{sec:related_work}

\paragraph{Models for Document Understanding.}
There are two main types of language models for document understanding: (1) OCR-dependent models, which use Optical Character Recognition (OCR) to extract text for processing, including the LayoutLM series~\cite{LayoutLMv1, LayoutLMv2, LayoutLMv3} and text-only LLMs~\cite{Llama3, Qwen2}; and (2) end-to-end models, which use a visual encoder to extract features from document images, integrating them with text for input into language model backbones. Most of document-related LVLMs fall into this category, such as GPT4o~\cite{GPT4o}, Gemini-1.5~\cite{Gemini15}, Claude-3.5-Sonnet~\cite{Claude35Sonnet}, mPLUG-DocOwl2~\cite{mPLUG-DocOwl2}, Qwen2-VL~\cite{Qwen2-VL}.

\paragraph{Methods for Long Document Understanding.}
To address the challenges of cross-page document understanding with excessively long context lengths, early approaches employed hierarchical encoding methods~\cite{MP-DocVQA,MPDoc-SAS,UVL-TG,MPDoc-RMT}. In these approaches, an independent encoder processes the OCR text and visual modal information for each page, which is then passed to a small language model decoder for cross-page contextual learning. However, this approach is limited by the redundancy in OCR inputs, which restricts the context length and leads to the accumulation of errors~\cite{LayoutLMv1,LayoutLMv2,DocFormer}. Recently, with the rise of multimodal large models, methods based on multimodal Retrieval-Augmented Generation (MM-RAG)~\cite{VisRAG,GRAM,MMVQA,CFRet-DVQA,CREAM,M3DocRAG} and end-to-end multi-page large models~\cite{MANTIS,LLaVA-Next-Interleave,Leopard} have emerged. These models leverage the world knowledge of large language models to enhance understanding. End-to-end approaches mitigate error accumulation by dynamically reducing the number of visual tokens across multiple pages/images and build large scale instruction turning dataset~\cite{MANTIS,LLaVA-Next-Interleave,Leopard,mPLUG-DocOwl2}, allowing for longer context lengths. Methods such as multi-page RAG facilitate dynamic interactions with OCR and other text information to remove redundant multimodal tokens.

\vspace{-0.4mm}
\paragraph{Benchmarks for Long Document Understanding.}
Multi-page or long documents place higher demands on the model's capabilities in cross-page understanding. Current multi-page document benchmarks, such as MP-DocVQA and DUDE, do not assess documents exceeding 20 pages. MMLongBench-Doc~\cite{MMLongBench-Doc} and M-Longdoc~\cite{M-Longdoc-arxiv} have been proposed to evaluate the understanding capabilities of longer documents, which have an average of 47.5 and 210.8 pages per document, respectively. Meanwhile, MMVQA~\cite{MMVQA} and MVQA~\cite{MVQA} are proposed to better evaluate the retrieval-based method. WebQuest~\cite{WebQuest} focuses on evaluating models' performance on text-rich web images.

\begin{figure*}[htbp]
    \centering
    \includegraphics[width=\textwidth]{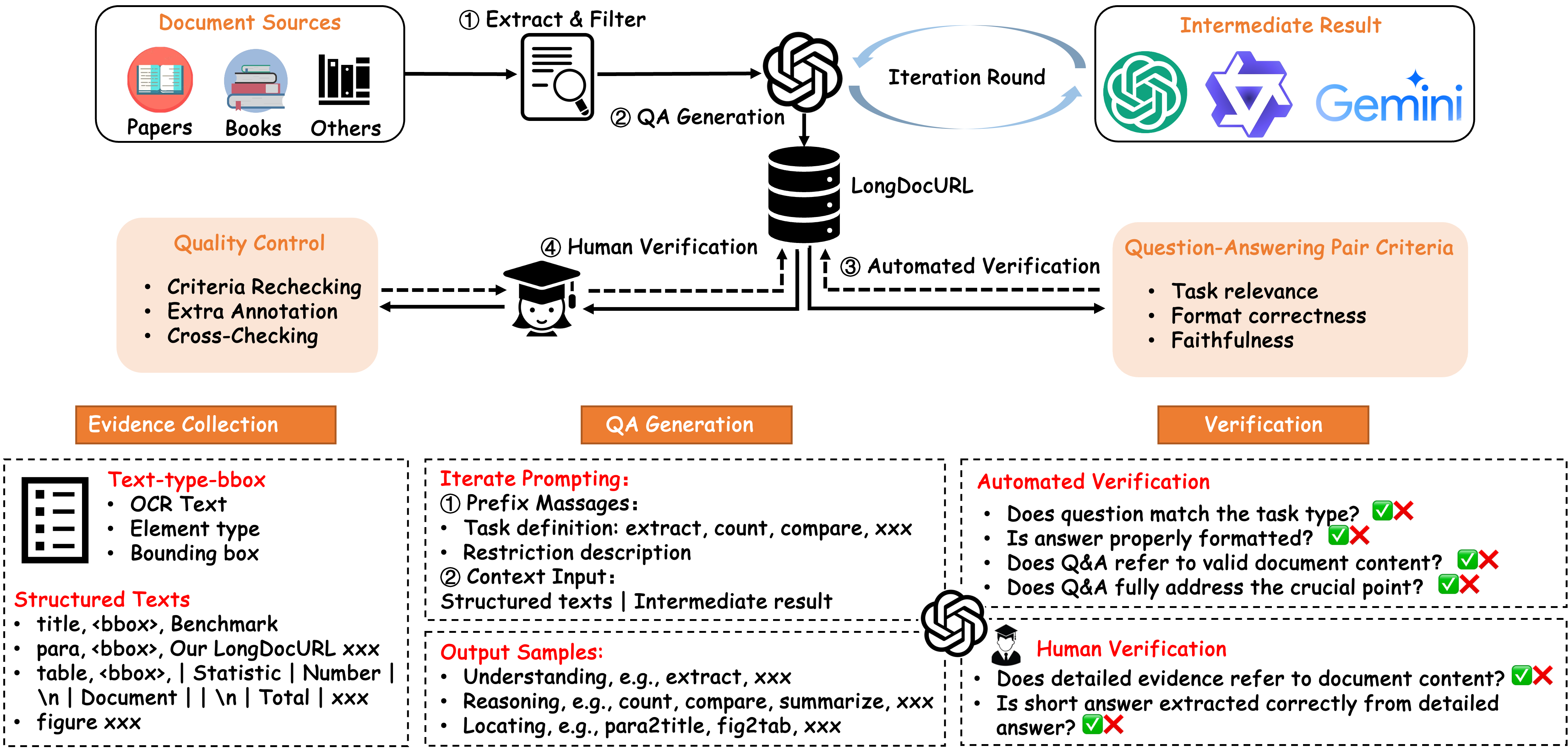} 
    \vspace{-0.3cm}
    \caption{Overview of our semi-automated construction pipeline. The pipeline comprises four modules: (a) \emph{Extract \& Filter}; (b) \emph{QA Generation}; (c) \emph{Automated Verification}; (d) \emph{Human Verification}.}
    \vspace{-0.2cm}
    \label{fig:pipeline}
\end{figure*}

\section{\benchName}
\label{sec:benchmark}

\subsection{Overview}
\label{subsec:overview}

\begin{figure}[t]
    \centering
    \includegraphics[width=0.4\textwidth]{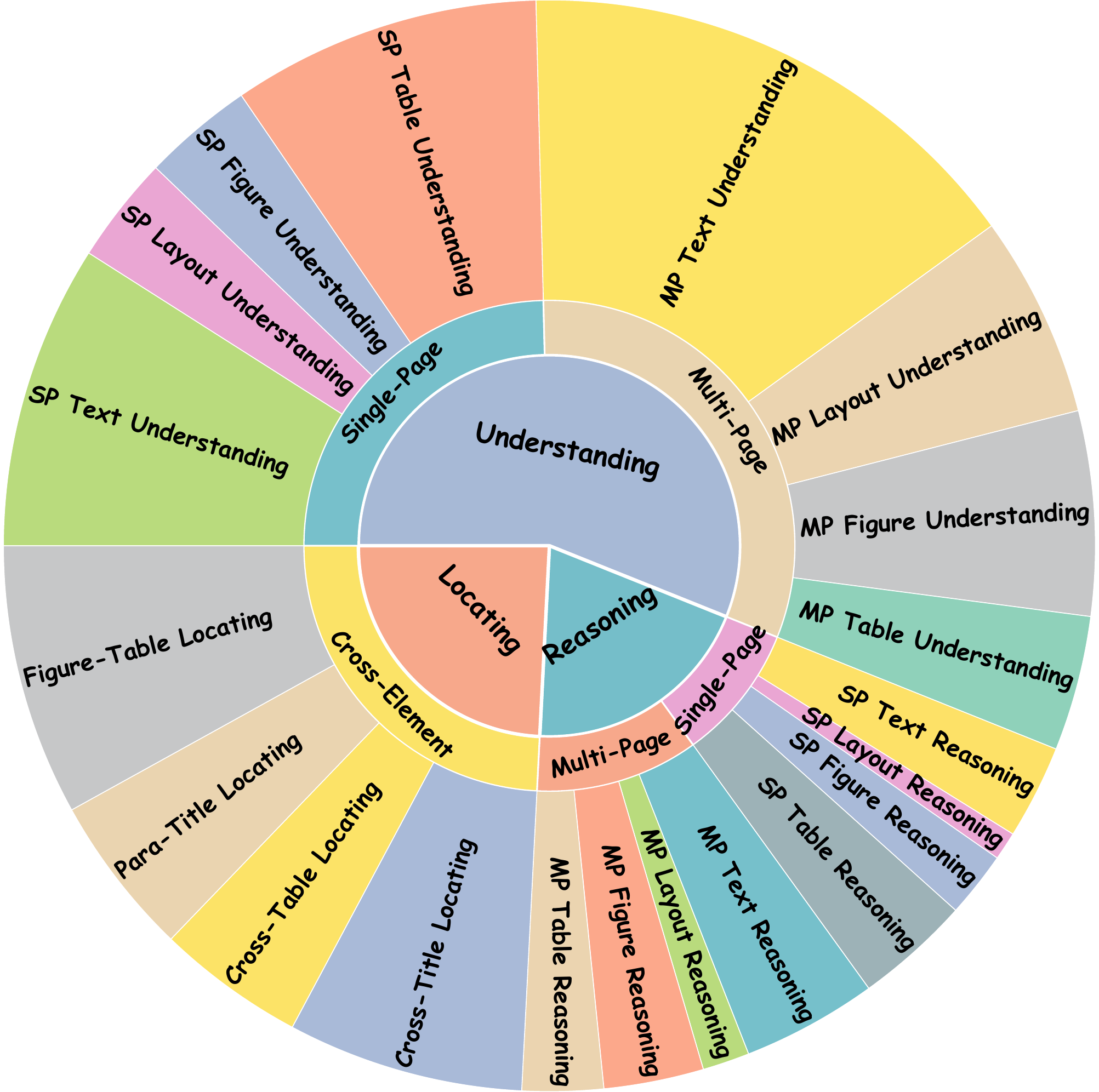} 
    \caption{Our \benchName~comprises 20 sub-tasks. \textbf{Inner}: divided by the primary task categories (Understanding, Reasoning, and \Locating). \textbf{Middle}: divided by the number of answer evidence pages (Single-Page, Multi-Page), and the number of types of evidence elements (Cross-Element). \textbf{Outer}: divided by the types of evidence elements (Text, Table, Figure, Layout).}
    \vspace{-0.5cm}
    \label{fig:task_classification_and_distribution}
\end{figure}

Firstly, each question-answering pair can be categorized by three primary tasks: \emph{Understanding}, \emph{Reasoning}, and \emph{\Locating}, as discussed in \Cref{sec:intro}. Secondly, we define four types of answer evidences based on element type: (1) \emph{Text}: pure texts, such as paragraph; (2) \emph{Layout}: text elements with special layout meaning (generalized text), such as title, header, footer, table name and figure name; (3) \emph{Figure}: including charts and general images. (4) \emph{Table}. In addition, each question-answering pair can be classified into \emph{single-page} or \emph{multi-page} based on the number of answer evidence pages and \emph{single-element} or \emph{cross-element} based on the number of types of evidence elements. Based on different primary tasks and answer evidences, we divide our dataset into 20 sub-tasks. As shown in \Cref{fig:task_classification_and_distribution}, for the \emph{Understanding} or \emph{Reasoning} task, we divide our dataset into 8 sub-tasks according to the number of answer evidence pages. Compared to the two tasks, we pay more attention to the interrelations
among different types of elements in the \emph{Locating} task and we build 4 sub-tasks based on the combination of different element types. Data examples are presented in \Cref{xsec:data_examples}.

\subsection{Q\&A Construction}

\subsubsection{Evidence Collection}
\label{subsubsec:evidence_collection}
To objectively evaluate LVLM long document question-answering comprehension ability, we first crawl 200k PDF-formatted documents from CommonCrawl\footnotemark[1] and filter them by page length(\ie, 50\textasciitilde 150) and language(\ie, English) to create a candidate set of approximately 3,000 documents. Then, we categorize these candidates by document type. Specifically, we randomly select 5\textasciitilde10 pages from a document, and prompt GPT-4o to classify its document type based on document content and layout. We finally retain 396 documents to construct our benchmark. These documents span eight types: \emph{research reports \& papers}, \emph{user manuals \& guides}, \emph{books \& e-books}, \emph{theses \& dissertations}, \emph{work \& project summaries}, \emph{presentation materials}, \emph{project proposals}, and \emph{meeting minutes \& summaries}, with an average of 85.6 pages per document.

\footnotetext[1]{https://corp.digitalcorpora.org/corpora/files/CC-MAIN-2021-31-PDF-UNTRUNCATED/}

Thereafter, we utilize both PyMuPDF\footnotemark[2] and Docmind\footnotemark[3] to parse the PDF and extract texts and layout information from the documents. For instance, the tables are converted into markdown format with Docmind. We organize the extracted results in the format of \emph{\textbf{text-type-bbox}} triples as a symbolic representation of elements: 1) \emph{\textbf{text}}: the recognized text of the elements; 2) \emph{\textbf{type}}: the element type. 3) \emph{\textbf{bbox}}: the bounding box of the element. Notably, we build the element triples at the region level, such as paragraph, table, chart, footnote and title, instead of the line level.

\footnotetext[2]{https://pymupdf.readthedocs.io}
\footnotetext[3]{https://www.aliyun.com/product/ai/docmind}

\subsubsection{Q\&A Generation}
\label{Data Generation}

Directly prompting LVLM to generate conversational question-answering pairs based on a single or multiple document images proves often ineffective, due to the inability to fully parse diverse elements present in the documents. Similar to LLaVA~\cite{LLaVA}, we adopt a two-stage pipeline: Initially, we parse our PDF-formatted documents and get the ``text-type-bbox'' triples discussed in \Cref{subsubsec:evidence_collection}. Subsequently, we design prompts to query LLMs/LVLMs in a multi-step round, and finally generate question-answering pairs. Specifically, as shown in \Cref{fig:pipeline}, we present the definition of each task and description of related restriction as a part of the prompts. Details can be found in \Cref{xsec:method_details}.

\subsection{Q\&A Verification}
\label{subsec:qa_verify}
\subsubsection{Automated Verification}
\label{subsubsec:auto_verify}
We design an automated method to verify the quality of synthesized question-answering pairs to identify and correct corresponding issues.
As shown in \Cref{fig:pipeline}, a qualified question-answering pair should be verified using these criteria: (1) \textbf{Task Relevance}. (2) \textbf{Format Correctness}. (3) \textbf{Faithfulness}.

\begin{table}[t]
    \centering
    \resizebox{0.45\textwidth}{!}{
            \begin{tabular}{lc}
            \toprule
            \textbf{Statistic} & \textbf{Number} \\
            
            \midrule
            \textbf{Document} \\
                ~- Total & 396 \\
                ~- Type & 8 \\
                ~- Avg. pages & 85.6 \\
              
            \midrule
            \textbf{Question \& Answer} \\
                ~- Total & 2,325 \\
                ~- Avg. question tokens & 35.5 \\
                ~- Max. question tokens & 277 \\
            \midrule
            \textcolor{gray}{Task Type} \\
                ~- Understanding & 1,243 (53.5\%) \\
                ~- Reasoning & 387 (16.6\%)  \\
                ~- \Locating & 695 (29.9\%) \\

            \midrule
            \textcolor{gray}{Type of Evidence Element} \\ 
                ~- Pure-text & 994 (42.8\%) \\
                ~- Layout & 779 (33.5\%) \\
                ~- Table & 556 (23.9\%) \\
                ~- Figure & 871 (37.5\%) \\

            \midrule
            \textcolor{gray}{Number of Evidence Pages} \\
                ~- Single-page question & 1,093 (47.0\%) \\
                ~- Multi-page questions & 1,230 (52.9\%)  \\
                ~- Unanswerable questions & 2 (0.1\%) \\

            \midrule
            \textcolor{gray}{Number of Evidence Element Types} \\
                ~- Single-element question & 1,463(62.9\%) \\
                ~- Cross-element question & 862(37.1\%) \\

            \midrule
            \textcolor{gray}{Answer Format} \\ 
                ~- String   & 941 (40.5\%) \\
                ~- Integer  & 431 (18.5\%) \\
                ~- Float & 185 (8.0\%) \\
                ~- List & 757 (32.6\%) \\
                ~- None & 11 (0.5\%) \\
            \bottomrule
        \end{tabular}}
        \captionof{table}{Document and Q\&A statistics based on different tasks and answer evidences.}
        \vspace{-0.5cm}
        \label{tab:statistics}
\end{table}

The verification results are utilized to classify samples: those that fail are marked as negative, and those that pass are marked as positive.
We analyze the verification results and observe approximately 75.2\% of the raw data in the Cross-Title \Locating~task are marked as negative samples, while the percentage is 19.6\% in the Cross-Table \Locating~task. Additional statistics data is presented in \Cref{tab:qa_numeric_verification}. In the next stage (\Cref{subsubsec:human_verify}), we present both the final result and texts of verification chains to the human annotator as reference information to guide further verification.

\subsubsection{Human Verification}
\label{subsubsec:human_verify}
There are two shortcomings of automated verification in \Cref{subsubsec:auto_verify}: (1) It only completes the classification of positive and negative samples, but does not recycle them, which causes waste. (2) When verifying the consistency between the question-answer pair and the document, only the text information is referenced, which may lost significant visual structure information compared to the source document. In the human verification stage, we focus on tasks that are challenging for automated verification, such as recovering negative samples and checking the consistency between the question-answering pair and the visual source information. The Human Verification module comprises the following parts: 

\begin{itemize}[leftmargin=*]
    \item \textbf{Criteria Rechecking.} The annotator reviews the intermediate process of machine verification to determine whether it is reasonable and whether the negative samples can be recovered by simply correcting the question-answer pair.
    \item \textbf{Extra Annotation.} Human annotators should oversee the question-answer pair process using the source visual document, not the parsed one. First, they need to verify if the evidence provided by the model aligns with the document's content, and then ensure that the final answer can be derived step by step from the intermediate process.
    \item \textbf{Cross-Checking.} After the annotators complete the first round of annotation, we require the annotators to cross-check each other's annotation results to improve the quality of annotation.
\end{itemize}

The labor resource allocation for data annotation is detailed in \Cref{xsec:labeling_labor}.

\subsection{Dataset Statistics}
\label{Dataset Statistics}

As shown in \Cref{tab:statistics} and \Cref{fig:document_distribution}, our \benchName~comprises 396 documents and 2,325 question-answering samples. The average pages per document range from 50 to 150. The cross-element locating task we designed contributes 37.1\% of the data, providing support for evaluating the model's cross-element interaction capabilities in long contexts. The dataset includes five types of answers, ensuring compatibility with automated evaluation while maintaining completeness and accuracy. In addition, we compare the characteristics of question set in our \benchName~with that in MMLongBench-Doc. The results are presented in \Cref{xsec:comparison_with_mmlongbenchdoc}.

\section{Experiments}
\label{sec: experiments}

\subsection{Evaluation Protocols}
\label{subsec: evaluation_protocols}
Allowing models to think freely can enhance performance but often results in lengthy responses. 
Current rule-based scorers are only effective with specific short formats, making it difficult to evaluate longer responses. Therefore, we need an \emph{answer extractor} (GPT4o) to convert detailed responses into concise ones. Following MATHVISTA~\cite{MATHVISTA} and MMLongBench-Doc~\cite{MMLongBench-Doc}, we implement a three-stage protocol: \emph{response generation}, \emph{answer extraction}, and \emph{score calculation}. In the \emph{score calculation} stage, we divide scores into five categories (\emph{Integer}, \emph{Float}, \emph{String}, \emph{List}, and \emph{None}). Different from MMLongBench-Doc, we adopt a softer and more reasonable scoring standard. See \Cref{xsec:experimental_details} for more details.

\subsection{Experimental Setup}
\label{subsec: experimental_setup}
\textbf{Models} We divide the experiments into two categories: \textbf{text-input} and \textbf{image-input}. For text-input configuration, document texts parsed by OCR engines are input into LLMs/LVLMs.
We conduct our experiments on both open-source and closed-source
models across 26 different configurations.

\noindent\textbf{Input Paradigm}
The documents in our \benchName~have a max of 150 pages, and most of models are unable to fully process the context to obtain the answer due to GPU memory or interface limitations. Therefore, similar to the \textbf{merge} method used in previous benchmarks~\cite{MMLongBench-Doc}, we design the \textbf{cut-off} paradigm for the evaluation of current models.

For LVLMs, we cut 30 continuous pages around the answer evidences from the original PDF-formatted document, and feed the converted images into the models. 
Details of selection rules are in \Cref{xsubsec:rules_of_images_selection}.
As for LLMs, we input the texts parsed by OCR engines, including PyMuPDF and Docmind.

\noindent\textbf{Other Configurations}
We assess the proprietary models using API resources, while the evaluation of the open-source models is conducted on H20 machines with 96G memory. To reduce variance, we set the temperature coefficient to 0.0 when generating free-form response and extracting short answer.

\begin{table*}[t]
    \centering
    \resizebox{\textwidth}{!}{
    \begin{tabular}{@{}l@{\hspace{1mm}}c|c@{\hspace{1mm}}c@{\hspace{1mm}}c@{\hspace{1mm}}c@{\hspace{1mm}}|c@{\hspace{1mm}}|c@{\hspace{1mm}}c@{\hspace{1mm}}c@{\hspace{1mm}}c@{\hspace{1mm}}|c@{\hspace{1mm}}|c@{\hspace{1mm}}c@{\hspace{1mm}}c@{\hspace{1mm}}c@{\hspace{1mm}}|c@{\hspace{1mm}}|c@{\hspace{1mm}}c@{\hspace{1mm}}c@{\hspace{1mm}}c@{\hspace{1mm}}|c@{\hspace{1mm}}|c@{\hspace{1mm}}c@{\hspace{1mm}}c@{\hspace{1mm}}c@{\hspace{1mm}}|c@{\hspace{1mm}}|c@{}}
    
        \toprule
        
        \multirow{4}{*}{\textbf{Model}} & \multirow{4}{*}{\textbf{Size}} & \multicolumn{10}{c|}{\textbf{Understanding}} & \multicolumn{10}{c|}{\textbf{Reasoning}} & \multicolumn{5}{c|}{\textbf{\Locating}} & \multirow{3}{*}{\textbf{Total}} \\
        
        & & \multicolumn{5}{c}{single-page} & \multicolumn{5}{c|}{multi-page} & \multicolumn{5}{c}{single-page} & \multicolumn{5}{c|}{multi-page} & \multicolumn{5}{c|}{cross-element} & \\
        
        & & TXT & LAY & FIG & TAB & \textcolor{blue}{all} & TXT & LAY & FIG & TAB & \textcolor{blue}{all} & TXT & LAY & FIG & TAB & \textcolor{blue}{all} & TXT & LAY & FIG & TAB & \textcolor{blue}{all} & CTi & CTa & PTi & FTa & \textcolor{blue}{all} & \\
        
        & & 259 & 91 & 94 & 263 & \textcolor{blue}{612} & 443 & 172 & 174 & 115 & \textcolor{blue}{631} & 40 & 12 & 28 & 98 & \textcolor{blue}{158} & 115 & 40 & 85 & 69 & \textcolor{blue}{229} & 201 & 126 & 137 & 231 & \textcolor{blue}{695} & 2325\\
        
        \midrule
        
        \multicolumn{28}{c}{\textit{OCR (PyMuPDF\footnotemark[4]) + Large Language Models (LLMs)} } \\
        
        \midrule
        
        {\textcolor{gray}{\textit{Open-source Models}}} \\
        
        LLaVA-Next-Interleave & 7B & 23.0 & 23.8 & 11.8 & 16.7 & \textcolor{blue}{19.5} & 36.2 & 32.3 & 25.0 & 29.2 & \textcolor{blue}{33.1} & 9.4 & 14.5 & 3.6 & 8.2 & \textcolor{blue}{8.7} & 27.3 & 15.0 & 20.8 & 23.3 & \textcolor{blue}{23.0} & 6.4 & 4.9 & 7.8 & 2.6 & \textcolor{blue}{5.8} & 18.7\\
        
        LLaVA-OneVision & 7B & 27.0 & 21.8 & 18.7 & 17.8 & \textcolor{blue}{22.7} & 43.1 & 37.6 & 33.6 & 33.4 & \textcolor{blue}{39.1} & 14.4 & 39.8 & 9.4 & 8.2 & \textcolor{blue}{10.4} & 31.9 & 25.0 & 29.0 & 20.5 & \textcolor{blue}{27.0} & 13.8 & 1.8 & 12.2 & 14.6 & \textcolor{blue}{11.2} & 23.3\\
        
        LLaVA-OneVision-Chat & 7B & 28.6 & 22.8 & 25.4 & 19.4 & \textcolor{blue}{24.6} & 43.0 & 36.3 & 35.4 & 31.0 & \textcolor{blue}{38.6} & 14.4 & 39.6 & 14.3 & 10.0 & \textcolor{blue}{12.4} & 31.2 & 20.0 & 28.0 & 22.4 & \textcolor{blue}{26.7} & 16.3 & 2.7 & 16.7 & 16.4 & \textcolor{blue}{14.0} & 24.6\\
        
        Qwen2-VL & 7B & 29.4 & 24.8 & 20.2 & 19.8 & \textcolor{blue}{24.5} & 42.3 & 36.6 & 33.6 & 33.3 & \textcolor{blue}{38.1} & 16.9 & 31.4 & 17.9 & 8.2 & \textcolor{blue}{12.5} & 32.0 & 22.5 & 33.8 & 23.7 & \textcolor{blue}{29.4} & 17.5 & 3.0 & 17.9 & 17.0 & \textcolor{blue}{14.9} & 25.0\\
        
        Qwen2.5-Instruct & 7B & 27.1 & 29.4 & 21.4 & 21.9 & \textcolor{blue}{24.7} & 36.5 & 32.9 & 33.8 & 28.0 & \textcolor{blue}{34.3} & 12.4 & 24.5 & 28.6 & 17.3 & \textcolor{blue}{18.3} & 32.0 & 32.5 & 29.2 & 24.2 & \textcolor{blue}{27.4} & 31.3 & 7.7 & 13.5 & 28.2 & \textcolor{blue}{20.5} & 25.9\\

        Qwen2.5-Instruct & 14B & 29.8 & 22.8 & 23.1 & 24.3 & \textcolor{blue}{26.4} & 38.4 & 34.6 & 37.4 & 33.5 & \textcolor{blue}{36.1} & 21.9 & 39.8 & 32.1 & 19.4 & \textcolor{blue}{21.4} & 29.3 & 35.0 & 22.1 & 22.7 & \textcolor{blue}{25.6} & 34.9 & 14.6 & 16.8 & 29.7 & \textcolor{blue}{24.2} & 27.9\\
        
        Qwen2.5-Instruct & 32B & 28.8 & 23.9 & 24.8 & 24.7 & \textcolor{blue}{26.6} & 34.9 & 32.6 & 31.3 & 36.0 & \textcolor{blue}{33.4} & 22.5 & 33.3 & 35.7 & 15.3 & \textcolor{blue}{20.3} & 31.1 & \colorbox{firstBest}{\textbf{40.0}} & 32.7 & 25.3 & \textcolor{blue}{29.4} & 33.5 & 23.1 & 28.5 & 31.2 & \textcolor{blue}{29.5} & 29.2\\
        
        Qwen2.5-Instruct & 72B & 31.9 & 30.3 & 23.3 & 25.3 & \textcolor{blue}{28.3} & 42.3 & 39.7 & 35.1 & 34.6 & \textcolor{blue}{39.4} & 22.5 & 25.0 & 35.7 & 15.3 & \textcolor{blue}{20.2} & \colorbox{firstBest}{\textbf{38.0}} & 39.9 & 32.6 & 26.7 & \textcolor{blue}{32.0} & 36.2 & 21.8 & 38.6 & 35.3 & \textcolor{blue}{34.2} & 32.9\\

        {\textcolor{gray}{\textit{Proprietary Models}}} \\
        
        Qwen-Max & - & 31.4 & 31.1 & 24.0 & 22.5 & \textcolor{blue}{27.1} & 40.0 & 35.3 & 33.5 & 34.8 & \textcolor{blue}{37.0} & 22.5 & 25.0 & 32.1 & 14.3 & \textcolor{blue}{19.0} & 31.9 & 37.5 & 31.5 & 22.9 & \textcolor{blue}{28.3} & 37.9 & 25.6 & 38.3 & 29.1 & \textcolor{blue}{34.0} & 31.4\\

        Gemini-1.5-Pro & - & 29.4 & 31.8 & 25.2 & 25.8 & \textcolor{blue}{27.8} & 41.3 & 40.3 & 35.5 & 32.0 & \textcolor{blue}{38.6} & 21.9 & 23.1 & 28.6 & 20.4 & \textcolor{blue}{21.4} & 31.9 & \colorbox{firstBest}{\textbf{40.0}} & \colorbox{firstBest}{\textbf{36.2}} & 24.3 & \textcolor{blue}{30.4} & 39.2 & 21.3 & 32.2 & 35.1 & \textcolor{blue}{32.8} & 32.0\\
        
        Qwen-VL-Max & - & \colorbox{firstBest}{\textbf{37.2}} & \colorbox{firstBest}{\textbf{37.0}} & 29.1 & 26.3 & \colorbox{firstBest}{\textcolor{blue}{\textbf{32.0}}} & \colorbox{firstBest}{\textbf{45.4}} & 44.1 & \colorbox{firstBest}{\textbf{39.9}} & \colorbox{firstBest}{\textbf{44.5}} & \colorbox{firstBest}{\textbf{\textcolor{blue}{43.3}}} & \colorbox{firstBest}{\textbf{28.7}} & \colorbox{firstBest}{\textbf{66.2}} & \colorbox{firstBest}{\textbf{45.8}} & 20.4 & \textcolor{blue}{26.8} & 37.1 & 37.5 & 33.8 & 27.1 & \textcolor{blue}{32.1} & 25.5 & 13.9 & 34.4 & 30.4 & \textcolor{blue}{27.3} & 33.3\\
        
        GPT-4o & - & 33.5 & 29.6 & 27.4 & 27.0 & \textcolor{blue}{30.4} & 42.6 & 42.7 & 38.0 & 33.5 & \textcolor{blue}{40.2} & 27.5 & 25.0 & 42.9 & 21.4 & \textcolor{blue}{25.9} & 33.7 & 37.5 & 31.5 & 25.6 & \textcolor{blue}{30.0} & \colorbox{firstBest}{\textbf{42.5}} & 24.1 & 40.8 & 35.6 & \textcolor{blue}{37.2} & 34.7\\
        
        O1-preview & - & 33.5 & 31.3 & \colorbox{firstBest}{\textbf{29.9}} & \colorbox{firstBest}{\textbf{29.6}} & \textcolor{blue}{31.1} & 41.7 & \colorbox{firstBest}{\textbf{45.0}} & 38.5 & 39.2 & \textcolor{blue}{40.1} & 22.5 & 33.3 & \colorbox{firstBest}{\textbf{45.8}} & \colorbox{firstBest}{\textbf{24.5}} & \colorbox{firstBest}{\textbf{\textcolor{blue}{27.1}}} & 34.9 & 37.5 & 29.2 & \colorbox{firstBest}{\textbf{38.6}} & \colorbox{firstBest}{\textbf{\textcolor{blue}{34.1}}} & 39.3 & \colorbox{firstBest}{\textbf{26.6}} & \colorbox{firstBest}{\textbf{44.9}} & \colorbox{firstBest}{\textbf{37.8}} & \colorbox{firstBest}{\textbf{\textcolor{blue}{38.6}}} & \colorbox{firstBest}{\textbf{35.8}}\\

        \midrule
        
        \multicolumn{28}{c}{\textit{Large Vision Language Models (LVLMs)} } \\
        
        \midrule
        
        {\textcolor{gray}{\textit{Open-source Models}}} \\
        
        InternLM-XC2.5 & 7B & 2.9 & 3.0 & 2.7 & 2.1 & \textcolor{blue}{2.7} & 5.5 & 5.3 & 3.2 & 2.7 & \textcolor{blue}{4.5} & 2.5 & 8.3 & 0.0 & 0.0 & \textcolor{blue}{0.6} & 3.5 & 2.5 & 1.2 & 1.3 & \textcolor{blue}{2.6} & 0.6 & 0.4 & 0.5 & 1.3 & \textcolor{blue}{0.7} & 2.4\\

        mPLUG-DocOwl2 & 7B & 6.2 & 6.9 & 4.6 & 3.5 & \textcolor{blue}{5.4} & 11.6 & 10.7 & 10.6 & 8.8 & \textcolor{blue}{9.9} & 2.5 & 8.3 & 0.0 & 1.0 & \textcolor{blue}{1.3} & 7.4 & 7.5 & 4.7 & 4.2 & \textcolor{blue}{5.5} & 2.4 & 0.0 & 1.8 & 2.4 & \textcolor{blue}{1.8} & 5.3 \\
        
        Pixtral & 12B & 10.1 & 8.2 & 6.7 & 4.6 & \textcolor{blue}{7.5} & 10.2 & 8.6 & 5.5 & 7.2 & \textcolor{blue}{8.4} & 2.7 & 8.9 & 0.0 & 0.0 & \textcolor{blue}{0.7} & 10.0 & 5.0 & 4.6 & 2.5 & \textcolor{blue}{5.8} & 3.7 & 1.3 & 1.9 & 2.2 & \textcolor{blue}{2.4} & 5.6\\
        
        Llama-3.2 & 11B & 11.9 & 9.9 & 13.1 & 8.1 & \textcolor{blue}{10.3} & 15.5 & 16.0 & 12.4 & 13.0 & \textcolor{blue}{15.3} & 2.5 & 8.3 & 10.7 & 6.1 & \textcolor{blue}{7.0} & 13.0 & 15.0 & 9.3 & 6.4 & \textcolor{blue}{11.0} & 3.0 & 1.4 & 3.8 & 1.6 & \textcolor{blue}{2.7} & 9.2\\

        LLaVA-Next-Interleave & 7B & 19.7 & 17.1 & 8.6 & 8.1 & \textcolor{blue}{14.0} & 30.3 & 25.4 & 19.7 & 16.6 & \textcolor{blue}{26.3} & 7.5 & 16.7 & 10.1 & 5.1 & \textcolor{blue}{6.9} & 21.4 & 12.5 & 19.9 & 8.5 & \textcolor{blue}{17.2} & 6.6 & 1.1 & 3.7 & 2.4 & \textcolor{blue}{3.8} & 14.1\\
        
        LLaVA-OneVision & 7B & 28.6 & 32.4 & 19.2 & 14.6 & \textcolor{blue}{22.3} & 36.7 & 37.1 & 28.1 & 23.5 & \textcolor{blue}{33.7} & 10.7 & 22.9 & 6.5 & 6.0 & \textcolor{blue}{7.6} & 26.4 & 20.0 & 24.0 & 12.2 & \textcolor{blue}{22.6} & 17.7 & 9.2 & 6.8 & 25.5 & \textcolor{blue}{14.1} & 22.0\\
        
        LLaVA-OneVision-Chat & 7B & 29.5 & 34.1 & 26.0 & 17.9 & \textcolor{blue}{24.8} & 36.9 & 38.4 & 34.5 & 26.9 & \textcolor{blue}{36.0} & 6.9 & 14.5 & 6.5 & 8.2 & \textcolor{blue}{8.0} & 31.8 & 22.5 & 29.3 & 16.5 & \textcolor{blue}{26.6} & 22.5 & 9.9 & 13.6 & 29.7 & \textcolor{blue}{18.7} & 25.0\\
        
        Qwen2-VL & 7B & 37.9 & 37.2 & 30.0 & 29.8 & \textcolor{blue}{33.4} & 43.8 & 38.0 & 37.8 & 33.5 & \textcolor{blue}{40.2} & 19.0 & 54.9 & 21.4 & 17.2 & \textcolor{blue}{19.3} & 33.2 & 22.5 & 34.5 & 16.5 & \textcolor{blue}{28.6} & 25.7 & 22.1 & 17.6 & 27.0 & \textcolor{blue}{22.6} & 30.6\\
        
        {\textcolor{gray}{\textit{Proprietary Models}}} \\

        Qwen-VL-Max & - & 61.1 & 54.5 & 58.2 & 53.7 & \textcolor{blue}{57.7} & 63.0 & 57.2 & 56.1 & 53.3 & \textcolor{blue}{60.0} & 44.1 & \colorbox{firstBest}{\textbf{66.6}} & 57.1 & 38.6 & \textcolor{blue}{44.0} & 45.8 & 42.5 & 38.5 & 41.0 & \textcolor{blue}{43.8} & 25.4 & 16.7 & 47.5 & 49.8 & \textcolor{blue}{36.0} & 49.5\\
        
        Gemini-1.5-Pro & - & 54.8 & 57.2 & 60.1 & 48.8 & \textcolor{blue}{52.8} & 61.4 & 56.9 & 54.2 & 51.8 & \textcolor{blue}{58.4} & 37.7 & 57.9 & 46.4 & 31.3 & \textcolor{blue}{35.9} & 53.4 & 52.5 & 47.8 & 40.4 & \textcolor{blue}{48.6} & 48.4 & 18.9 & 46.7 & 68.1 & \textcolor{blue}{46.4} & 50.9\\
        
        GPT-4o & - & \colorbox{firstBest}{\textbf{71.1}} & \colorbox{firstBest}{\textbf{67.5}} & \colorbox{firstBest}{\textbf{68.5}} & \colorbox{firstBest}{\textbf{66.4}} & \colorbox{firstBest}{\textbf{\textcolor{blue}{68.6}}} & \colorbox{firstBest}{\textbf{70.1}} & \colorbox{firstBest}{\textbf{67.4}} & \colorbox{firstBest}{\textbf{69.1}} & \colorbox{firstBest}{\textbf{63.8}} & \colorbox{firstBest}{\textbf{\textcolor{blue}{68.6}}} & \colorbox{firstBest}{\textbf{56.7}} & \colorbox{firstBest}{\textbf{66.6}} & \colorbox{firstBest}{\textbf{70.8}} & \colorbox{firstBest}{\textbf{49.5}} & \colorbox{firstBest}{\textbf{\textcolor{blue}{55.7}}} & \colorbox{firstBest}{\textbf{67.3}} & \colorbox{firstBest}{\textbf{55.0}} & \colorbox{firstBest}{\textbf{61.7}} & \colorbox{firstBest}{\textbf{61.1}} & \colorbox{firstBest}{\textbf{\textcolor{blue}{62.8}}} & 50.7 & \colorbox{firstBest}{\textbf{40.1}} & \colorbox{firstBest}{\textbf{66.8}} & \colorbox{firstBest}{\textbf{78.5}} & \colorbox{firstBest}{\textbf{\textcolor{blue}{59.6}}} & \colorbox{firstBest}{\textbf{64.5}}\\

        \midrule
        \multicolumn{28}{c}{\textit{Human Baseline} } \\
        \midrule
        Human Experts & - & - & - & - & - & \textcolor{blue}{-} & - & - & - & - & \textcolor{blue}{87.2} & - & - & - & - & \textcolor{blue}{-} & - & - & - & - & \textcolor{blue}{81.3} & - & - & - & - & \textcolor{blue}{81.6} & 84.8 \\
        
        \bottomrule
    \end{tabular}}
    \caption{Generalized accuracy scores(0\textasciitilde1) on \benchName. There are 3 types of tasks: (U)nderstanding, (R)easoning, and (L)ocating. There are 4 types of evidence elements: pure text(TXT), layout(LAY), chart \& image(FIG), and table(TAB). There are 3 types of evidence pages/elements: single-page(SP), multi-page(MP), and cross-element(CE). Cross-element means at least two element types in the evidence(\eg, chart and table). CTi: Cross-Title, CTa: Cross-Table, PTi: Para-Title, FTa: Figure-Table. The highest scores among models in each section are highlighted in \colorbox{firstBest}{\textbf{green}}. \textit{Human baselines} are obtained from 102 uniformly distributed samples.}
    \label{tab:main_results_1}
    \vspace{-5mm}
\end{table*}

\subsection{Main Results}
\label{subsec: main_results}
As shown in \Cref{tab:main_results_1} and \Cref{tab:main_results_2}, we calculate generalized accuracy scores to assess model capabilities. Regarding LVLMs, we draw the following conclusions: (1) \textbf{Highest Scoring Model}: Only GPT-4o meets the passing standard and scores highest at 64.5, indicating that our \benchName~presents significant challenges for current models. (2) \textbf{Comparison of Open-source and Closed-source models}: Proprietary models demonstrate better overall performance compared to open-source models. Among open-source models, only Qwen2-VL (score 30.6) and LLaVA-OneVision (scores 22.0 and 25.0) exceed a score of 20, while other models with fewer than 13B parameters fall below this threshold.

To compare the performance of models using text input versus image input, we included the O1-preview and Qwen2.5 series. The experimental results show that the overall scores of LLMs significantly lower than LVLMs, with the top LLM score trailing the top LVLM score by about 30 points. This gap is mainly because important document information is lost when parsed into plain text using PyMuPDF. Our dataset features numerous table-related and chart-related Q\&A pairs, and the loss of structural information hinders LLMs’ ability to extract critical evidence. These results highlights our \benchName~ as a benchmark for assessing the document structure parsing capability of LVLMs.

\subsubsection{Fine-Grained Analysis}
\label{subsubsec:multi-dimensional_analysis}

\begin{figure*}[htbp]
    \centering
    \includegraphics[width=\textwidth]{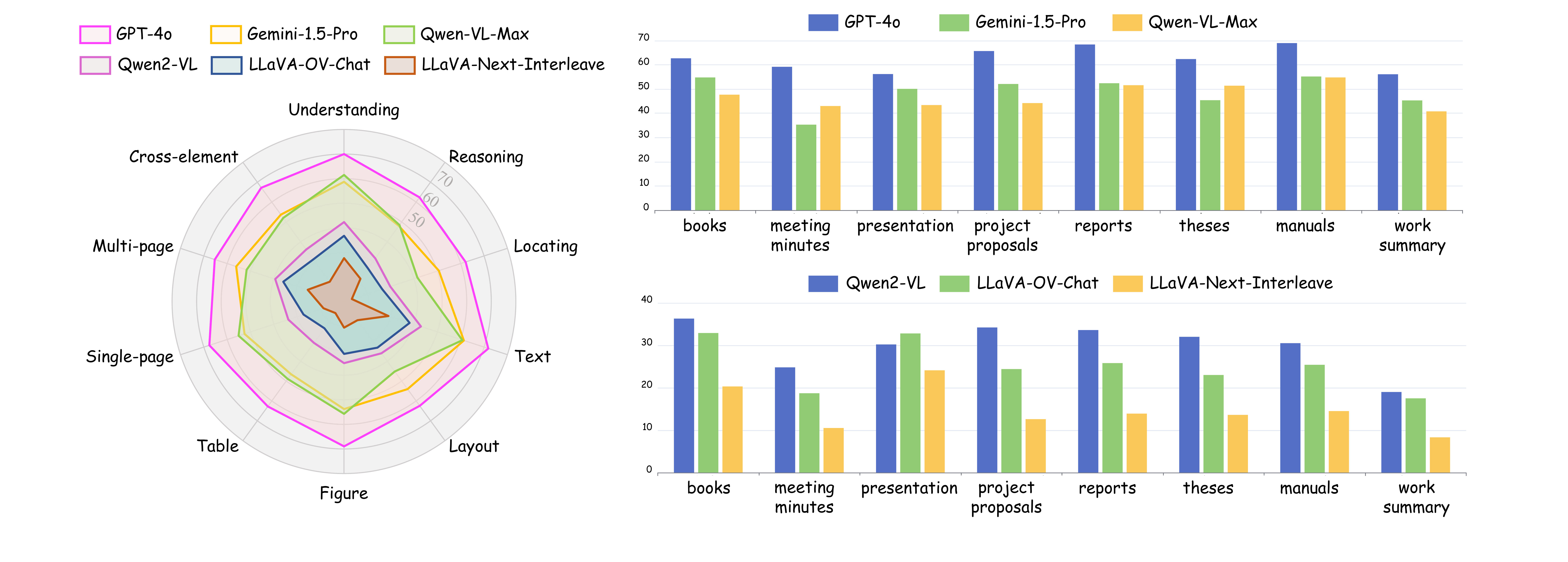} 
    \vspace{-0.5cm}
    \caption{Fine-grained Results. We choose 3 proprietary and 3 open-source models to conduct further analysis based on (left) \textbf{task types}, \textbf{document elements}, \textbf{evidence pages}, and (right) \textbf{document sources}.}
    \vspace{-0.2cm}
    \label{fig:fine_grained_results}
\end{figure*}

We include more fine-grained results in \Cref{tab:main_results_2} and \Cref{fig:fine_grained_results}, based on document sources, task categories, document elements, and evidence pages.

\noindent\textbf{Document Sources} 
As shown in \Cref{fig:fine_grained_results}, the models perform better on \emph{books}, \emph{reports}, \emph{manuals}, and \emph{project proposals}, likely due to their simpler layout, which facilitates key information extraction. Conversely, the models struggle with less common documents like \emph{meeting minutes} and \emph{work summary}, which have limited data.

\noindent\textbf{Task Type} 
Observations reveal: (1) Proprietary LVLMs perform comparably on reasoning and \locating~tasks, but image-to-text conversion impacts reasoning capabilities more severely. For instance, switching to text input, GPT-4o's reasoning scores drop by 31.6 points versus 22.4 points for \locating. (2) Strong models are balanced in reasoning and \locating, whereas weaker models perform poorly on \locating, suggesting a training focus on capabilities of understanding and reasoning over spatial and logical relationships in \locating~tasks.

\noindent\textbf{Document Elements} 
Models score highest on \emph{Text} questions and lowest on \emph{Table} ones, highlighting deficiencies in document structure parsing. \emph{Figure} and \emph{Layout} question types yield similar scores. Scores for cross-element tasks fall between single-page and multi-page QA, closely aligning with the overall assessment.

\noindent\textbf{Single-page vs Multi-page} 
Single-page QA accuracy is lower than multi-page QA. This reveals that answers for some questions can be gathered from multiple pages, thereby reducing the difficulty. However, models like GPT-4o and Qwen-VL-Max show lower accuracy on multi-page QA, revealing a contradiction where their scores on \locating~tasks in multi-page QA are lower, thus skewing overall performance.

\begin{table}[htbp]
    \centering
    \resizebox{0.5\textwidth}{!}{
    \begin{tabular}{@{}lc|cc|ccc@{}}
        \toprule
        \multirow{2}{*}{\textbf{Model}} & \multirow{2}{*}{\textbf{Size}} & \multicolumn{2}{c|}{\textbf{Image-input}} & \multicolumn{2}{c}{\textbf{Text-input}} \\
        & & cut-off & merge & pymupdf & docmind \\
        \midrule
        \textcolor{gray}{\textit{Open-source Models}} \\
        Qwen2-VL & 7B & 30.7 & 7.6\footnotemark[4] & 24.4 & 45.3 \\
        LLaVA-Next-Interleave & 7B & 12.6 & 10.9 & 16.9 & 29.7 \\
        LLaVA-OneVision-Chat & 7B & 24.1 & 14.0 & 23.0 & 39.1 \\
        
        \textcolor{gray}{\textit{Proprietary Models}} \\
        Gemini-1.5-Pro & - & 48.1 & -\footnotemark[5] & 34.0 & 64.8 \\
        GPT-4o & - & 64.4 & 44.9 & 36.5 & 66.2 \\
        O1-preview & - & - & - & 38.0 & 63.4 \\
        
        \bottomrule
    \end{tabular}
    }
    \caption{Comparison among different input paradigms on a subset of 20\% data.}
    \label{tab:results_in_different_input_formats}
    \vspace{-6mm}
\end{table}

\footnotetext[4]{363 out of 465 Q\&A pairs meet OOM problems with Qwen2-VL. We calculate normalized score on the size of 465.}
\footnotetext[5]{Not completed due to resource limitations.}

\subsection{Ablation of Input Paradigms}
\label{subsec: ablation_of_input_paradigms}

To explore the optimal input format in long document question-answering, we conduct ablation experiments across two image-input and two text-input paradigms. The image-input paradigms include: (1) \textbf{cut-off}, following the configuration detailed in \Cref{subsec: experimental_setup}, and (2) \textbf{merge}, where document images are combined from raw document lengths (50\textasciitilde150) into 20\textasciitilde30 new images. Further details can be found in \Cref{xsubsec:rules_of_images_merging}. 

We note that the table structure information significantly degrades when parsed by PyMuPDF, while the markdown-format table texts parsed by Docmind retain greater structural integrity. To assess the impact of structural information loss on model performance, we conducted experiments with two input types: (1) \textbf{text-input-docmind}, utilizing texts parsed by Docmind, and (2) \textbf{text-input-pymupdf}, utilizing texts parsed by PyMuPDF. The analysis of the results presented in \Cref{tab:results_in_different_input_formats} led us to the following conclusions:

\noindent\textbf{Text-input vs. Image-input:} 
The scores in the \emph{cut-off} paradigm are higher than that in the \emph{text-input-pymupdf} paradigm, but lower than that in the \emph{text-input-docmind} paradigm, indicating that this method can effectively extract table structure information, but it can be improved further. 

\noindent\textbf{Cut-off vs. Merge:}
The \emph{merge} method preserves a greater number of context tokens by concatenating multiple images, while the \emph{cut-off} method succeeds in acquiring prior information by shortening the context window. Experimental results suggest that the cut-off may yield better problem-solving capabilities than merging, providing insights for the future construction of multimodal Retrieval-Augmented Generation (RAG) systems.

\noindent\textbf{Impact of Structural Information:}
For proprietary models, performance utilizing Docmind is at least 25 points higher than that with PyMuPDF, while the disparity is 15 points for open-source models. The absence of table structure information significantly hampers the performance of both open-source and proprietary models.

\section{Error Analysis and Model Insights}
\label{sec:error_analysis_and_model_insights}

\begin{figure}[t]
    \centering
    \includegraphics[width=0.42\textwidth]{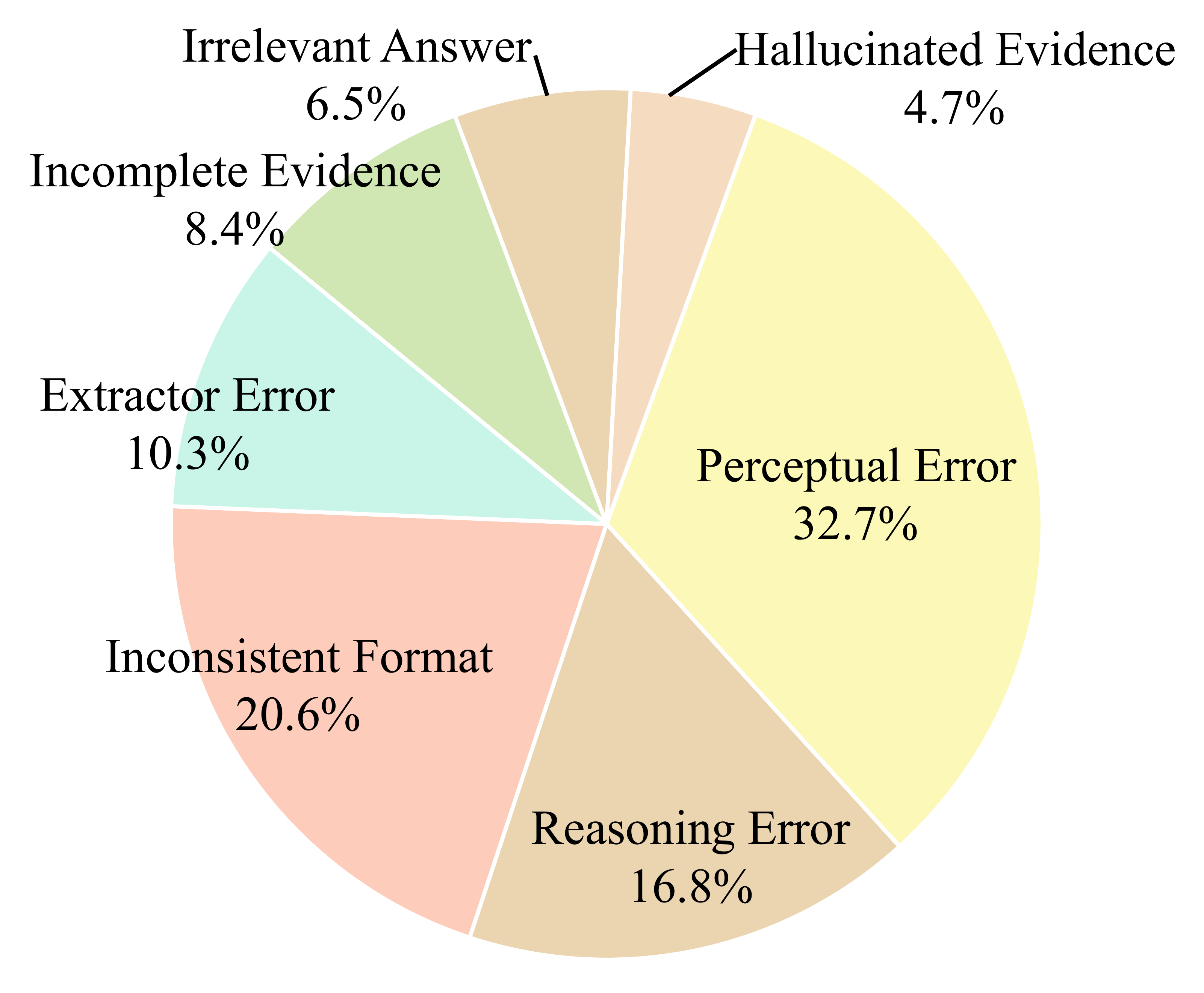}
    \vspace{-0.1cm}
    \caption{Error distribution over 97 annotated GPT-4o errors.}
    \label{fig:error_distribution}
    \vspace{-0.5cm}
\end{figure}

\paragraph{Error Analysis}
We randomly sample 97 erroneous records and conduct a error analysis of the best-performing model, GPT-4o. We classify the errors into 7 types (See \Cref{xsec:error_analysis} for detailed descriptions and qualitative examples) and show their distribution in \Cref{fig:error_distribution}.

\begin{figure}[t]
    \vspace{0.3cm}
    \centering
    \includegraphics[width=0.50\textwidth]{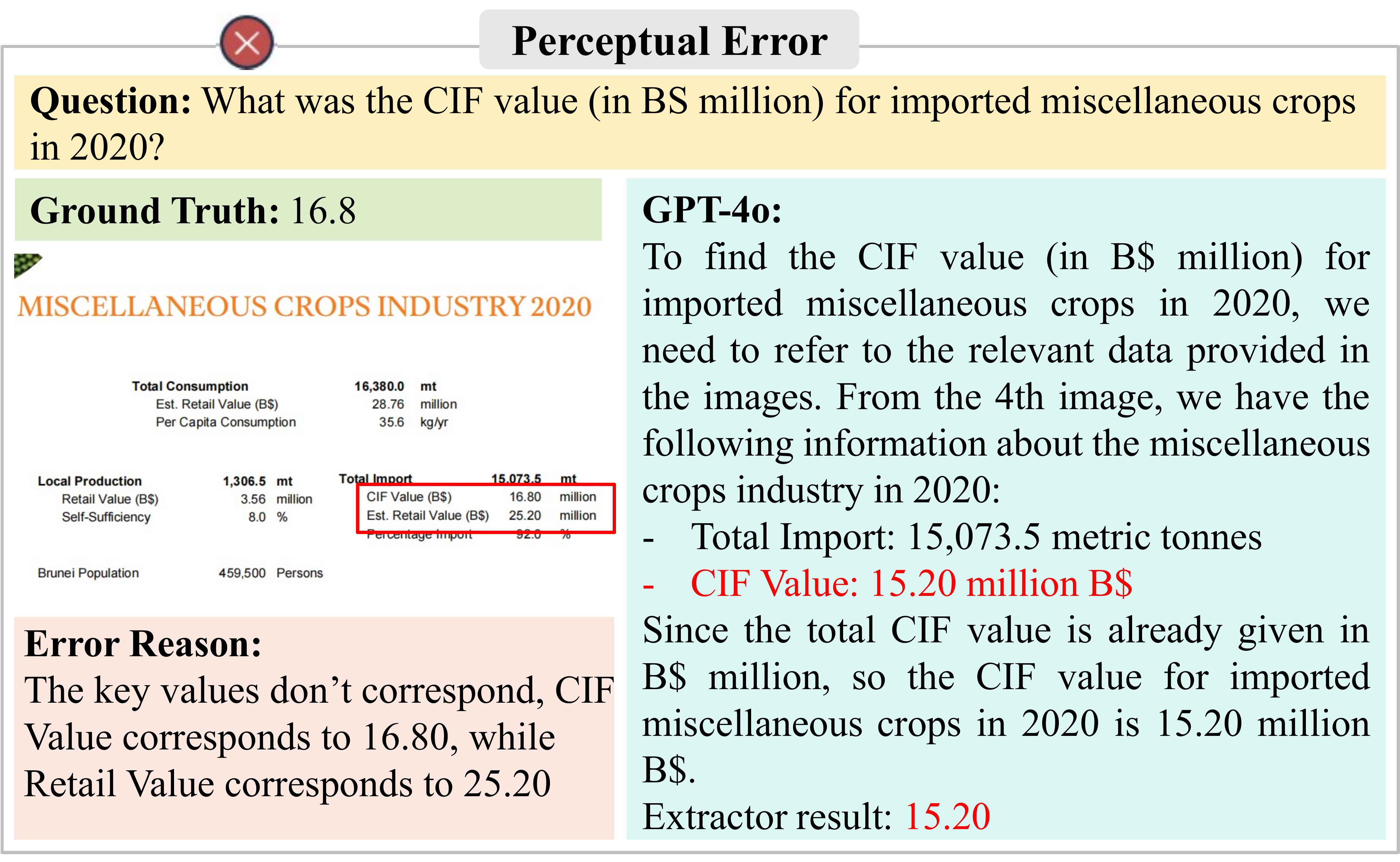}
    \caption{A basic perception error, with the error highlighted in red.}
    \label{fig:perception_error_example_before_appendix}
    \vspace{-0.4cm}
\end{figure}

\emph{Perceptual errors} constitute the most prevalent category (32.7\%), arising from inaccuracies in recognizing, parsing, or counting document elements such as tables, figures, and formulas. Typical manifestations include incorrect heading hierarchies and mismatched figure-text correspondences, indicating that complex document structures and OCR performance remain primary bottlenecks for LVLMs. An example of perceptual errors in LongDocURL is shown in \Cref{fig:perception_error_example_before_appendix}. \emph{Reasoning errors} represent the third-largest category (16.8\%), occurring during calculation, comparison, or summarization even when relevant evidence is correctly identified. A significant proportion of errors (20.6\%) involves \emph{inconsistent formats}, highlighting the inflexibility of rule-based evaluation. For instance, a response of "50212000" for "\$50.2 million" was incorrectly scored as entirely wrong. The remaining error categories (\emph{Hallucinated Evidence}, \emph{Irrelevant Answer}, \emph{Incomplete Evidence}, and \emph{Extractor Error}) collectively accounted for 29.9\% of cases.

\paragraph{Model Insights}
We analyze open-source models from two key aspects: \emph{dataset adaptation} and \emph{training strategy}. First, Qwen2-VL (LVLM) demonstrates superior performance over its LLM variant in \Cref{tab:main_results_1}, attributable to its extensive multimodal training that strengthens both comprehension and generation capabilities across modalities. Second, LLaVA-OneVision-Chat outperforms its base model (LLaVA-OneVision), as it employ Direct Preference Optimization (DPO) and human feedback to boost generalization and reasoning in long-document tasks. Additionally, we observe frequent failures caused by poor layout analysis and table/chart parsing. Incorporating layout parsing into continual pretraining or fine-tuning could mitigate these errors. Alternatively, a multi-stage response framework---where models first parse document structure before reasoning---may improve performance.

\vspace{-2mm}
\section{Conclusion}
In this study, we address the limitations of existing document benchmarks. We propose \benchName, which includes 20 capabilities across 3 tasks, 3 evidence modes, and 4 document elements. A semi-automated pipeline generated over 2,300 high-quality question-answer pairs, covering more than 33,000 pages of documents. Subsequently, we conducted a comprehensive evaluation of 26 different parameter amounts of both open-source and closed-source models, revealing potential gaps in document understanding.

\section*{Limitations}
From the perspective of dataset source, the types of documents discussed in this paper are still limited. A broader range of data sources would provide richer document layouts and element information, which would be more beneficial for evaluation. Moreover, the dataset can be further expanded by the automated construction pipeline. On the other hand, designing better model structures and training processes to improve performance on \benchName will be more important. However, this may have gone beyond the scope of this paper.

\section*{Acknowledgments}
This work has been supported by the National Natural Science Foundation of China (NSFC) grant U23B2029 and Zhongguancun Academy Project No.20240102.

\bibliography{main}
\clearpage
\appendix

\section{More Statistics of \benchName}
\label{xsec:more_statistics}
\begin{figure*}
    \centering
    \includegraphics[width=0.9\textwidth]{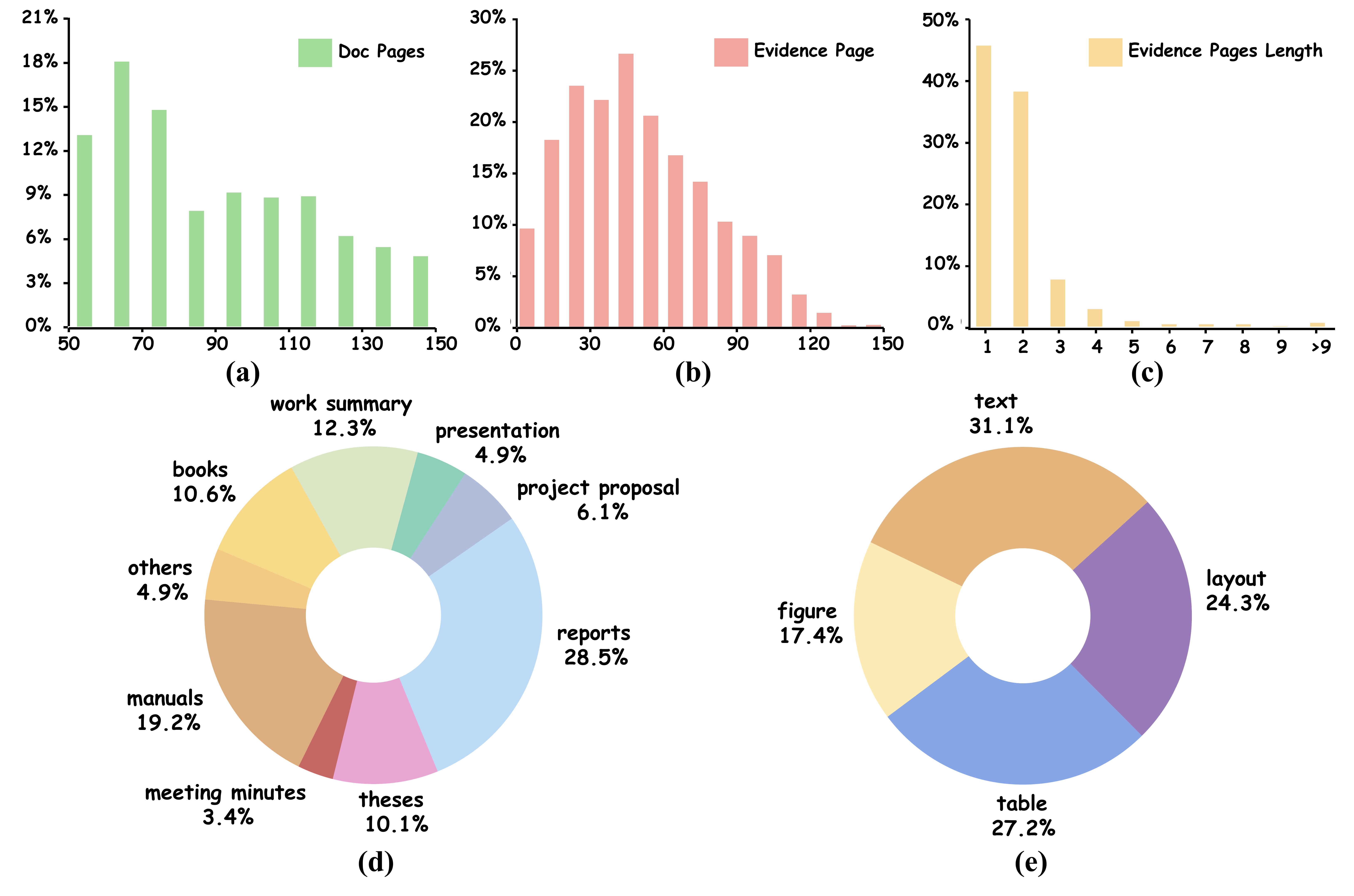} 
    \caption{The statistical analysis of our dataset about the distribution characteristics across (a) \textbf{document pages}, (b) \textbf{answer evidence page}, (c) \textbf{evidence pages length}, (d) \textbf{document sources}, and (e) \textbf{evidence element types}.}
    \label{fig:document_distribution}
\end{figure*}

\Cref{fig:document_distribution} illustrate our dataset distribution characteristics across document pages, answer evidence page, evidence pages length, document sources, and evidence element types.

\section{Method Details}
\label{xsec:method_details}

\subsection{QA Construction}
\label{xsubsec:details_of_qa_construction}

\subsubsection{Prompt Template for QA Generation}

\begin{tcolorbox}[colback=gray!5]
\small
\textbf{[System]}\\ 
You are an expert in document question-answering dialogue synthesis. Please complete the following instructions based on the given text. The response must be true and accurate, and no additional content should be output. \\

\textbf{[Task Description]}\\
Your task is \textcolor{blue}{<detailed\_task\_description>}\\

\textbf{[Restriction]} \\
Ensure questions and answers are suitable and correct.

Only include questions that have definite answers, that is:
\begin{itemize}[leftmargin=*]
    \item one can see the content in the image that the question asks about and can answer confidently;
    \item one can determine confidently from the image that it is not in the image, don't ask any question that cannot be answered confidently.
\end{itemize}

Provide detailed evidence description first and then give final short answers. Use examples or reasoning steps to support your content. You can include multiple paragraphs if necessary.

\textcolor{blue}{<other\_restriction\_description>} \\

\textbf{[Response Examples]}\\
\textcolor{blue}{<few\_qa\_examples>} \\

\textbf{[Context Input]}\\
\textcolor{blue}{<structured\_text> | <previous\_response>} \\

\label{text:prompt_template_qa_construction}
\end{tcolorbox}

Notes: \textcolor{blue}{<structured\_text>} refers to \textbf{text-type-bbox} triple processed by Docmind engine, which is discussed in detail in \Cref{subsubsec:evidence_collection}. \textcolor{blue}{<previous\_response>} refers to possible intermediate result(\eg, summary sentences in our Para-Title \Locating~task. Specifically, we prompt LVLMs to generate summary sentences of one or multiple paragraphs under certain section titles first, and then utilize these summaries and title names to construct question-answering data.)

\subsubsection{Prompt Template for QA Verification}

\begin{tcolorbox}[colback=gray!5]
\small
\textbf{[System]}\\ 
You are an expert in document question-answering verification. Please complete the following instructions based on the given text. The response must be true and accurate, and no additional content should be output. \\

\textbf{[Task description \& Verification criteria]} \\ 
Your task is to ensure the quality of question-answer pairs in the context provided. you need to follow the steps outlined below to systematically evaluate each pair's effectiveness and accuracy. Implement this process diligently to maintain high standards across the batch of QA pairs. \\

1. Question type check \\

Does the question match the task description: \textcolor{blue}{<task\_description>}

Make sure the question meets the required task context. \\

2. Formatting and Presentation \\

Is answer properly formatted?

Ensure the answer uses a list format to store the title content. \\

3. Relevance Check \\
Does the question relate directly to document content rely on the context provided, the answer accurately reflect the information in the document?

Ensure the question is formulated based on information explicitly stated or implied in the document. The question should not introduce concepts unrelated to the document's content. 

Validate that the answer references specific data or statements from the document. Avoid including extraneous information not supported by the document. \\

4. Clarity and Precision \\
Is the question clear and unambiguous? And is the answer concise and precise?
Ensure the language is straightforward and easily understandable, avoid complex phrasing that may confuse the reader.

The intention of the question and answer pair must be clear and direct, avoiding verbosity and unnecessary detail.

Ensure the answer fully addresses the question without omitting crucial information. \\

5. Consistency and Coherence \\
Check for logical flow and coherence, ensuring the question aligns seamlessly with the document's narrative or arguments. Verify that the answer does not contradict or misrepresent other sections of the document.

By practicing this process, you can confirm whether the quality of the question-answer pairs meets the requirements. If all the above conditions are met, please output yes, otherwise output no. \\

Generated QA: \textcolor{blue}{<qa>} \\

Input Context: \textcolor{blue}{<evidence\_context>}

\label{text:prompt_template_qa_verification}
\end{tcolorbox}

\subsubsection{Statistics of QA Verification}
\label{xsubsubsec:statistics_of_qa_verification}
\begin{table}[htbp]
    \centering
    \scriptsize
    \vspace{-1.0mm}
    \resizebox{0.4\textwidth}{!}{
        \begin{tabular}{@{}l@{\hspace{15mm}}c@{\hspace{15mm}}c@{}}
        \toprule
        \textbf{Verify}  & \textbf{U + R}       & \textbf{L} \\
        
        \midrule
        \textbf{Before} & 2857 & 1520 \\
        \textbf{After} & 1630 & 695 \\                               
        &57.1\%↓   &45.7\%↓ \\            
        \bottomrule
        \end{tabular}}
    \caption{Statistics of changes in the amount of data before and after verification.}
    \label{tab:qa_numeric_verification}
    \vspace{-5mm}
\end{table}

\section{Comparison with MMLongBench-Doc}
\label{xsec:comparison_with_mmlongbenchdoc}
\begin{figure}[htbp]
    \centering
    \vspace{-0.5mm}
    \includegraphics[width=0.4\textwidth]{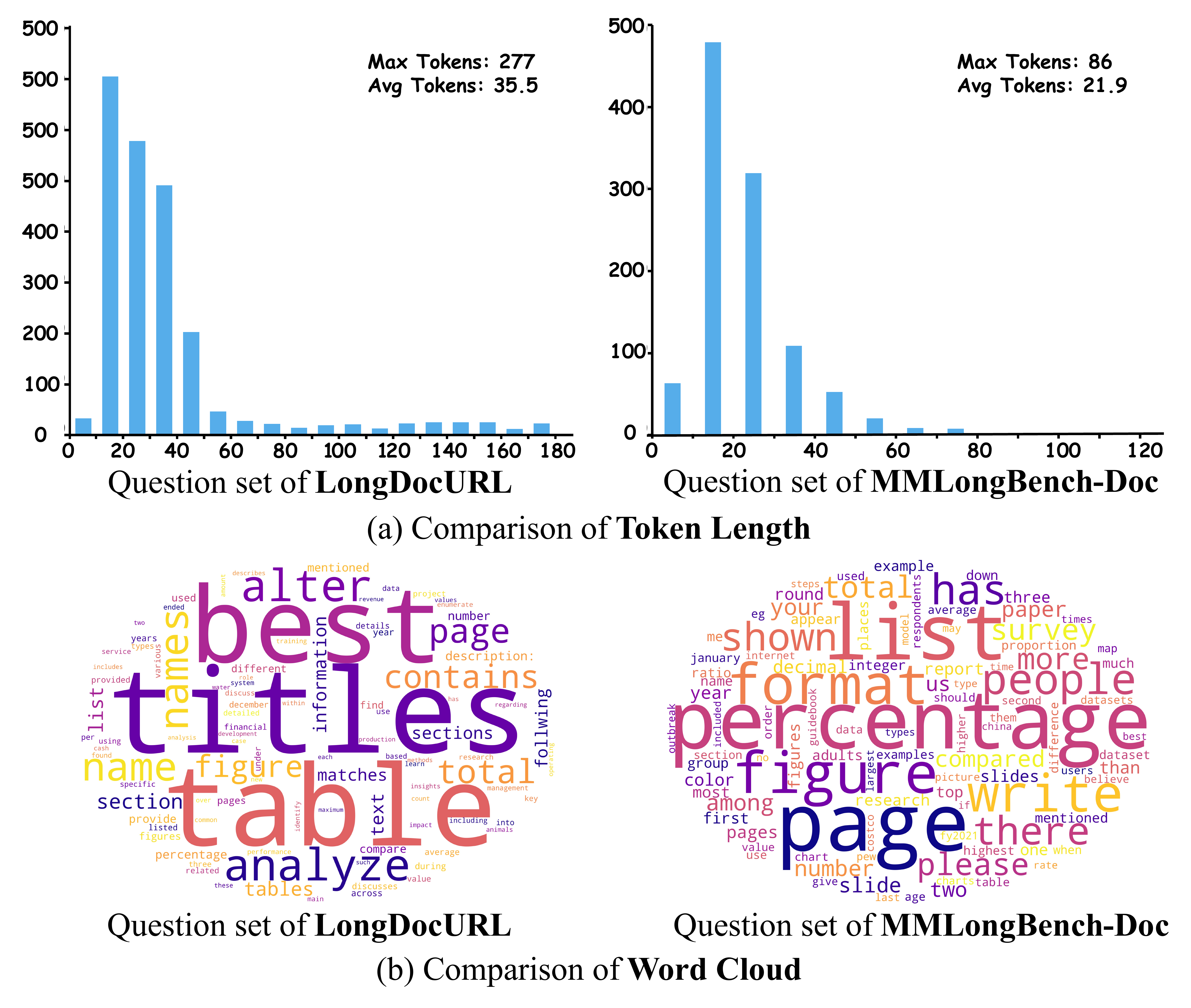} 
    \caption{The dataset attributes comparison between our \benchName~and MMLongBench-Doc.}
    \label{fig:comparison_with_mmlongbenchdoc}
    \vspace{-0.5mm}
\end{figure}

\section{Experimental Details}
\label{xsec:experimental_details}

\subsection{Merging Rules of Images Input}
\label{xsubsec:rules_of_images_merging}
As discussed in \Cref{subsec: ablation_of_input_paradigms}, we design a \textbf{merge} paradigm for the evaluation of current LLMs/LVLMs.

\noindent\texttt{\#Columns\_merged = 2 if total\_pages/30 <= 4 else 3}

\begin{table}[htbp]
    \centering
    \resizebox{0.5\textwidth}{!}{
    \begin{tabular}{l|c|c}
        \toprule
        \textbf{Total\_pages} & \textbf{\#Columns\_merged} & \textbf{\#Images\_merged} \\
        \midrule
        50<x<=60 & 2 & x/2 \\
        60<x<=90 & 2 & x/3 \\
        90<x<=120 & 2 & x/4 \\
        120<x<=150 & 3 & x/5 \\
        \bottomrule
    \end{tabular}
    }
    \caption{Merging rules in the ablation experiment of input paradigms. \textbf{\#Columns\_merged}: the number of columns in the sub-image array in the merged image; \textbf{\#Images\_merged}: the number of new images after merged.}
    \label{tab:merging_rules}
    \vspace{-5mm}
\end{table}

\subsection{Selection Rules of Images Input}
\label{xsubsec:rules_of_images_selection}
As discussed in \Cref{subsec: experimental_setup}, we design a \textbf{cut-off} paradigm for the evaluation of current LLMs/LVLMs. We provide pseudo-code below to express the selection rules.

\begin{tcolorbox}[colback=gray!5,left=0.8mm,right=0.8mm]
\small
\textcolor{gray}{\# All page ids mentioned are based on the order in the raw document.} \\
\textcolor{gray}{\# context\_start\_end: page start/end id of context in qa generation} \\
\textcolor{gray}{\# num\_of\_images\_input: the number of input images, 30 images in cut-off paradigm} \\
\textcolor{gray}{\# total\_pages: total pages in the raw document} \\
\textcolor{gray}{\# img\_start, img\_end: input page start/end id} \\
\raggedright
\texttt{raw\_start\_page, raw\_end\_page = context\_start\_end} \\

\texttt{raw\_pages\_len = raw\_end\_page - raw\_start\_page} \\

\texttt{img\_start = max(0, raw\_start\_page -}  \\
\texttt{\quad \quad (num\_of\_images\_input - raw\_pages\_len)//2} \\

\texttt{img\_end = img\_start + num\_of\_images\_input} \\

\texttt{if img\_end >= total\_pages:} \\

\texttt{\quad \quad img\_end = total\_pages} \\

\texttt{\quad \quad img\_start = max(0, } \\
\texttt{\quad \quad img\_end - num\_of\_images\_input)} \\

\texttt{return img\_start, img\_end}
        
\label{text:selection_rules}
\end{tcolorbox}

\subsection{Prompt for Response Generation}
\label{xsubsec:prompt_of_response_generation}


\begin{tcolorbox}[colback=gray!5]
\small
\textcolor{blue}{<document\_images>} \\

You are an expert in visual document question-answering, please answer our questions based
on the given doc images.\\

Following is our question:\\
<question\_start>\textcolor{blue}{<question>}</question\_end>

\label{text:prompt_for_response_generation}
\end{tcolorbox}

\subsection{Prompt for Answer Extraction}
\label{xsubsec:prompt_of_answer_extract}

Prompt for answer extraction is displayed in \Cref{fig:answer_extraction}. Based on the template given in MMLongBench-Doc~\cite{MMLongBench-Doc}, we make some modifications, which are marked in \textcolor{blue}{blue}.

\begin{figure*}[htbp]
\begin{tcolorbox}[title=The answer extracting prompt used to make long response concise during evaluation,colback=gray!5,colframe=gray!75!gray]
{\footnotesize


\vspace{0.1cm}
\textbf{[Task Description]}

Given the question and analysis, you are tasked to extract answers with required formats from the free-form analysis. 

\vspace{0.1cm}

Your extracted answers should be one of the following formats: (1) Integer, (2) Float, (3) String and (4) List. If you find the analysis the question can not be answered from the given documents, type "Not answerable". Exception: If the analysis only tells you that it can not read/understand the images or documents, type "Fail to answer". 

\vspace{0.1cm}

\textcolor{blue}{Please use <concise\_answer> and <\/concise\_answer> tokens at the start and end of the extracted answer. For example, if the extracted answer is number 3, the format is <concise\_answer>3<\/concise\_answer>.}

\vspace{0.1cm}

\textcolor{blue}{Please use <answer\_format> and <\/answer\_format> tokens at the start and end of the answer format. For example, if the answer format is List, the format is <answer\_format>List<\/answer\_format>.}

\vspace{0.1cm}

Please make your response as concise as possible. Also note that your response should be formatted as below:

\vspace{0.2cm}

\textcolor{blue}{Extracted answer: <concise\_answer>[answer]<\/concise\_answer>}

\textcolor{blue}{Answer format: <answer\_format>[answer format]<\/answer\_format>}

\vspace{0.2cm}

\textbf{[Few Shots]}

Please read the following example, then extract the answer from the model response and type it at the end of the prompt. 

\vspace{0.2cm}

--- Example 1:

\textbf{Question}: List the primary questions asked about the services in this report.

\textbf{Analysis}: The primary questions asked about the services in the report for The Limes Residential Home are: 

1. Is the service safe?

2. Is the service effective?

3. Is the service caring?

4. Is the service responsive?

5. Is the service well-led? 

\textbf{Extracted answer}: <concise\_answer>['Is the servife safe?', 'Is the service effective', 'Is the serve caring?', 'Is the service responsive?', 'Is the service well-led?']<\/concise\_answer>

\textbf{Answer format}: <answer\_format>List<\/answer\_format>

\vspace{0.2cm}

--- Example 2:

\textbf{Question}: How many regulations of the HSCA 2008 are breached in all according to this report?

\textbf{Analysis}: According to the report, the provider breached 10 Health and Social Care Act 2008 (Regulated Activities) Regulations in total. Here are the specifics:

1. Regulation 13: Safeguarding service users from abuse and improper treatment

2. Regulation 12: Safe care and treatment

3. Regulation 18: Staffing

4. Regulation 11: Need for consent

5. Regulation 10: Dignity and respect

6. Regulation 9: Person-centred care

7. Regulation 17: Good governance

8. Regulation 18 (CQC Registration Regulations 2009): Notification of other incidents

9. Regulation 18: Failure to maintain an accurate and up-to-date care plan

10. Regulation 11: Failure to implement the Mental Capacity Act 2005 code of practice effectively

These breaches involve issues concerning staffing, safeguarding, medicines management, dignity and respect, consent, care planning, governance, and failure to notify the CQC of incidents.

\textbf{Extracted answer}: <concise\_answer>10<\/concise\_answer>

\textbf{Answer format}: <answer\_format>Integer<\/answer\_format>

\vspace{0.2cm}

--- Example 3:

\textbf{Question}: According to the survey that is the percentage of Chinese who are paying more or about the same attention to politics after Trump’s election?

\textbf{Analysis}: The survey provided does not specify the percentage of Chinese individuals specifically who are paying more or about the same attention to politics after Trump’s election. The report focuses primarily on American demographics and does not include specific details about the Chinese population in relation to this question. If you need information about a different demographic or a summary of the findings from the American demographic, I can certainly help with that!

\textbf{Extracted answer}: <concise\_answer>Not answerable<\/concise\_answer>

\textbf{Answer format}: <answer\_format>None<\/answer\_format>

\vspace{0.2cm}

--- Example 4:

\textbf{Question}: How many quotations from male respondent over 50 years old are included in this report?

\textbf{Analysis}: The image you’ve provided appears to be a screenshot of a document with multiple charts. However, the text is too small and blurry to read accurately. If you can provide a clearer image or more context, I might be able to help you with your question.

\textbf{Extracted answer}: <concise\_answer>Fail to answer<\/concise\_answer>

\textbf{Answer format}: <answer\_format>None<\/answer\_format>

}
\end{tcolorbox}
\vspace{-0.2cm}
\caption{Prompt for answer extraction during evaluation.}
\label{fig:answer_extraction}
\end{figure*}

\subsection{Scoring Rules}
\label{xsubsec:scoring_rules_details}

Following MATHVISTA~\cite{MATHVISTA}, we evaluate the model's responses by scoring the extracted answers against the reference answers. Following MMLongBench-Doc~\cite{MMLongBench-Doc}, our scorer is rule-based and employs different strategies according to the format of the reference answer. We detail its rules as below:

\noindent\textbf{String:} We firstly use a series of regular expressions to determine whether the answers require exact matching (\eg, telephone numbers, email addresses, website addresses, file names, times, dates, \etc.) If an exact match is needed, we perform a straightforward string comparison and score the answer either 0 or 1. Otherwise, we calculate the ANLS (Average Normalized Levenshtein Similarity) with a pre-defined threshold ($\tau = 0.5$).

\noindent\textbf{Integer:} We perform an exact match comparison and score the answer either 0 or 1.

\noindent\textbf{Float:} We view the prediction and reference answers as equal if they fall within a 1\% relative tolerance. 

\noindent\textbf{List:} Compared with MMLongBench-Doc, we adopt a relatively soft rule for scoring answers in list format: (1) If the prediction does not have the same number of elements as the reference, it incurs a length-dependent penalty instead of receiving a score of 0, which we think more reasonable. (2) The score of models on single element of the reference list is the highest one among the scores which are calculated and combined between the element and each one of the prediction list. Compared with MMLongBench-Doc, we assume that the sorting positions of the two lists are not always one-to-one corresponding, allowing more errors, and our rules are gentler and more tolerant. We use pseudo-code below to express the scoring rules in MMLongBench-Doc and our \benchName, respectively. The element-wise scoring strategies is determined by the formats of elements (\ie, string, integer or float).

\begin{tcolorbox}[colback=gray!5]
\small
\textcolor{gray}{\# MMLongBench-Doc} \\
pred\_list, ref\_list = sorted(pred\_list), sorted(ref\_list) \\
Score(pred\_list, ref\_list) = min([score(pred, ref) for pred, ref in zip(pred\_list, ref\_list)]) \\

\vspace{0.5em}
\textcolor{gray}{\# LongDocURL} \\
pred\_list, ref\_list = sorted(pred\_list), sorted(ref\_list) \\
greedy\_scores\_list = [ \\
\setlength{\parindent}{2em} 
\hspace{1em} max([score(pred, ref) for pred in pred\_list]) for ref in ref\_list \\
] \\
Score(ref\_list, pred\_list) = \\ 
\hspace{1em} sum(greedy\_scores\_list) / len(ref\_list) * min(1, len(ref\_list) / len(pred\_list)) ** 0.5 \\

\label{text:scoring_rules_for_list}
\end{tcolorbox}

The part of the rule description that differs from MMLongBench-Doc is mainly in the \textbf{List} section.

\subsection{Fine-Grained Evaluation Results}
\label{xsubsec:appendix_multi_dimensional_results}
Detailed results are presented in \Cref{tab:main_results_2}. 

\noindent Related analysis is in \Cref{subsubsec:multi-dimensional_analysis}.

\begin{table*}[htbp]
    \centering
    \resizebox{\textwidth}{!}{
    \begin{tabular}{lc|ccc|cccc|ccc|c}
        \toprule
        \multirow{3}{*}{\textbf{Model}} & \multirow{3}{*}{\textbf{Size}} & \multicolumn{3}{c|}{\textbf{Task Type}} & \multicolumn{4}{c|}{\textbf{Evidence Element}} & \multicolumn{3}{c|}{\textbf{Page/Element}}& \multirow{2}{*}{\textbf{Total}} \\

        & & U & R & L & TXT & LAY & FIG & TAB & SP & MP & CE & \\
        & & 1243 & 387 & 695 & 994 & 779 & 556 & 871 & 1093 & 1230 & 862 & 2325 \\
        
        \midrule
        
        \multicolumn{13}{c}{\textit{OCR (PyMuPDF\footnotemark[4]) + Large Language Models (LLMs)} } \\
        
        \midrule
        
        \textcolor{gray}{\textit{Open-source Models}} \\
        LLaVA-Next-Interleave & 7B & 26.4 & 17.2 & 5.8 & 26.0 & 13.8 & 15.4 & 14.0 & 14.4 & 22.6 & 16.0 & 18.7 \\
        LLaVA-Next-Interleave-DPO & 7B & 28.7 & 16.4 & 6.4 & 27.4 & 15.7 & 17.7 & 14.2 & 15.1 & 24.3 & 17.5 & 20.0 \\
        LLaVA-OneVision & 7B & 31.0 & 20.2 & 11.2 & 32.6 & 19.2 & 22.1 & 15.1 & 17.5 & 28.5 & 21.0 & 23.3 \\
        LLaVA-OneVision-Chat & 7B & 31.7 & 20.9 & 14.0 & 33.1 & 19.8 & 25.3 & 16.6 & 20.0 & 28.8 & 22.2 & 24.6 \\
        Qwen2-VL & 7B & 31.4 & 22.5 & 14.9 & 33.2 & 20.6 & 25.0 & 17.2 & 20.7 & 28.7 & 22.6 & 25.0 \\
        Qwen2.5-Instruct & 7B & 29.6 & 23.7 & 20.5 & 31.4 & 27.0 & 24.3 & 18.2 & 21.5 & 29.7 & 24.2 & 25.9 \\
        Qwen2.5-Instruct & 14B & 31.3 & 23.9 & 24.2 & 33.3 & 29.3 & 24.8 & 21.6 & 23.5 & 31.8 & 26.5 & 27.9 \\
        Qwen2.5-Instruct & 32B & 30.0 & 25.7 & 29.5 & 31.9 & 30.4 & 28.9 & 25.8 & 26.4 & 31.6 & 29.6 & 29.2 \\
        Qwen2.5-Instruct & 72B & 33.9 & 27.2 & 34.2 & 37.4 & 33.8 & 34.0 & 28.6 & 29.3 & 36.0 & 35.3 & 32.9 \\

        \textcolor{gray}{\textit{Proprietary Models}} \\
        Qwen-Max & - & 32.1 & 24.5 & 34.0 & 34.6 & 32.8 & 32.6 & 27.7 & 28.3 & 34.1 & 33.2 & 31.4 \\
        Gemini-1.5-Pro & - & 33.3 & 26.7 & 32.8 & 35.5 & 34.7 & 32.4 & 26.9 & 28.3 & 35.4 & 33.1 & 32.0 \\
        Qwen-VL-Max & - & \colorbox{firstBest}{\textbf{37.7}} & 30.0 & 27.3 & \colorbox{firstBest}{\textbf{39.6}} & 31.2 & 35.1 & 27.9 & 31.1 & 35.2 & 34.2 & 33.3 \\
        GPT-4o & - & 35.4 & \colorbox{firstBest}{\textbf{28.3}} & 37.2 & 37.6 & 36.3 & 36.3 & 30.2 & 31.9 & 37.2 & 35.8 & 34.7 \\
        O1-preview & - & 35.6 & 31.2 & \colorbox{firstBest}{\textbf{38.6}} & 37.4 & \colorbox{firstBest}{\textbf{37.1}} & \colorbox{firstBest}{\textbf{37.3}} & \colorbox{firstBest}{\textbf{34.6}} & \colorbox{firstBest}{\textbf{33.5}} & \colorbox{firstBest}{\textbf{37.8}} & \colorbox{firstBest}{\textbf{38.6}} & \colorbox{firstBest}{\textbf{35.8}} \\

        \midrule
        
        \multicolumn{13}{c}{\textit{Large Vision Language Models (LVLMs)} } \\
        
        \midrule
        
        \textcolor{gray}{\textit{Open-source Models}} \\
        InternLM-XC2.5 & 7B & 3.6 & 1.8 & 0.7 & 3.9 & 2.2 & 1.9 & 1.2 & 1.9 & 2.9 & 2.3 & 2.4 \\
        mPLUG-DocOwl2 & 7B & 7.7 & 3.8 & 1.8 & 8.1 & 4.7 & 5.5 & 3.1 & 3.7 & 6.7 & 6.1 & 5.3 \\
        Pixtral & 12B & 7.9 & 3.7 & 2.4 & 8.8 & 4.8 & 4.3 & 3.1 & 4.8 & 6.2 & 5.3 & 5.6 \\
        Llama-3.2 & 11B & 12.9 & 9.4 & 2.7 & 11.8 & 6.9 & 8.7 & 6.3 & 7.9 & 10.3 & 6.8 & 9.2 \\
        LLaVA-Next-Interleave & 7B & 20.2 & 13.0 & 3.8 & 21.8 & 10.8 & 12.1 & 6.7 & 10.0 & 17.8 & 11.3 & 14.1 \\
        LLaVA-Next-Interleave-DPO & 7B & 21.6 & 13.9 & 7.6 & 22.5 & 13.9 & 15.4 & 8.7 & 12.1 & 19.8 & 13.5 & 16.2 \\
        LLaVA-OneVision & 7B & 28.1 & 16.5 & 14.1 & 30.8 & 23.9 & 17.9 & 11.6 & 16.5 & 26.8 & 20.8 & 22.0 \\
        LLaVA-OneVision-Chat & 7B & 30.5 & 19.0 & 18.7 & 32.2 & 26.5 & 24.4 & 15.4 & 19.8 & 29.7 & 24.2 & 25.0 \\
        Qwen2-VL & 7B & 36.9 & 24.8 & 22.6 & 37.7 & 29.7 & 28.6 & 23.7 & 27.2 & 33.6 & 29.9 & 30.6 \\
        
        \textcolor{gray}{\textit{Proprietary Models}} \\
        Qwen-VL-Max & - & 58.8 & 43.9 & 36.0 & 58.0 & 40.2 & 52.3 & 44.6 & 51.6 & 47.6 & 48.0 & 49.5 \\
        Gemini-1.5-Pro & - & 55.7 & 43.4 & 46.4 & 58.7 & 50.4 & 50.0 & 41.8 & 48.7 & 52.8 & 49.9 & 50.9 \\
        GPT-4o & - & \colorbox{firstBest}{\textbf{68.6}} & \colorbox{firstBest}{\textbf{59.9}} & \colorbox{firstBest}{\textbf{59.6}} & \colorbox{firstBest}{\textbf{70.7}} & \colorbox{firstBest}{\textbf{60.0}} & \colorbox{firstBest}{\textbf{67.4}} & \colorbox{firstBest}{\textbf{60.3}} & \colorbox{firstBest}{\textbf{65.8}} & \colorbox{firstBest}{\textbf{63.2}} & \colorbox{firstBest}{\textbf{65.4}} & \colorbox{firstBest}{\textbf{64.5}} \\

        \midrule
        \multicolumn{13}{c}{\textit{Human Baseline} } \\
        \midrule
        Human Experts & - & - & - & - & - & - & - & - & - & - & - & 84.8 \\
        
        \bottomrule
    \end{tabular}
    }
    \caption{Generalized accuracy scores(0\textasciitilde1) on \benchName. There are 3 types of tasks: (U)nderstanding, (R)easoning, and (L)ocalization. There are 4 types of evidence elements: pure text(TXT), layout(LAY), chart \& image(FIG), and table(TAB). There are 3 types of evidence pages/elements: single-page(SP), multi-page(MP), and cross-element(CE). Cross-element means at least two element types in the evidence(\eg, chart and table). The highest scores among models in each section are highlighted in \colorbox{firstBest}{\textbf{green}}.}
    \label{tab:main_results_2}
    \vspace{-5mm}
\end{table*}

\section{Case Study}
\label{xsec:case_study}

\begin{figure*}[!htbp]
    \centering
    \includegraphics[width=\textwidth]{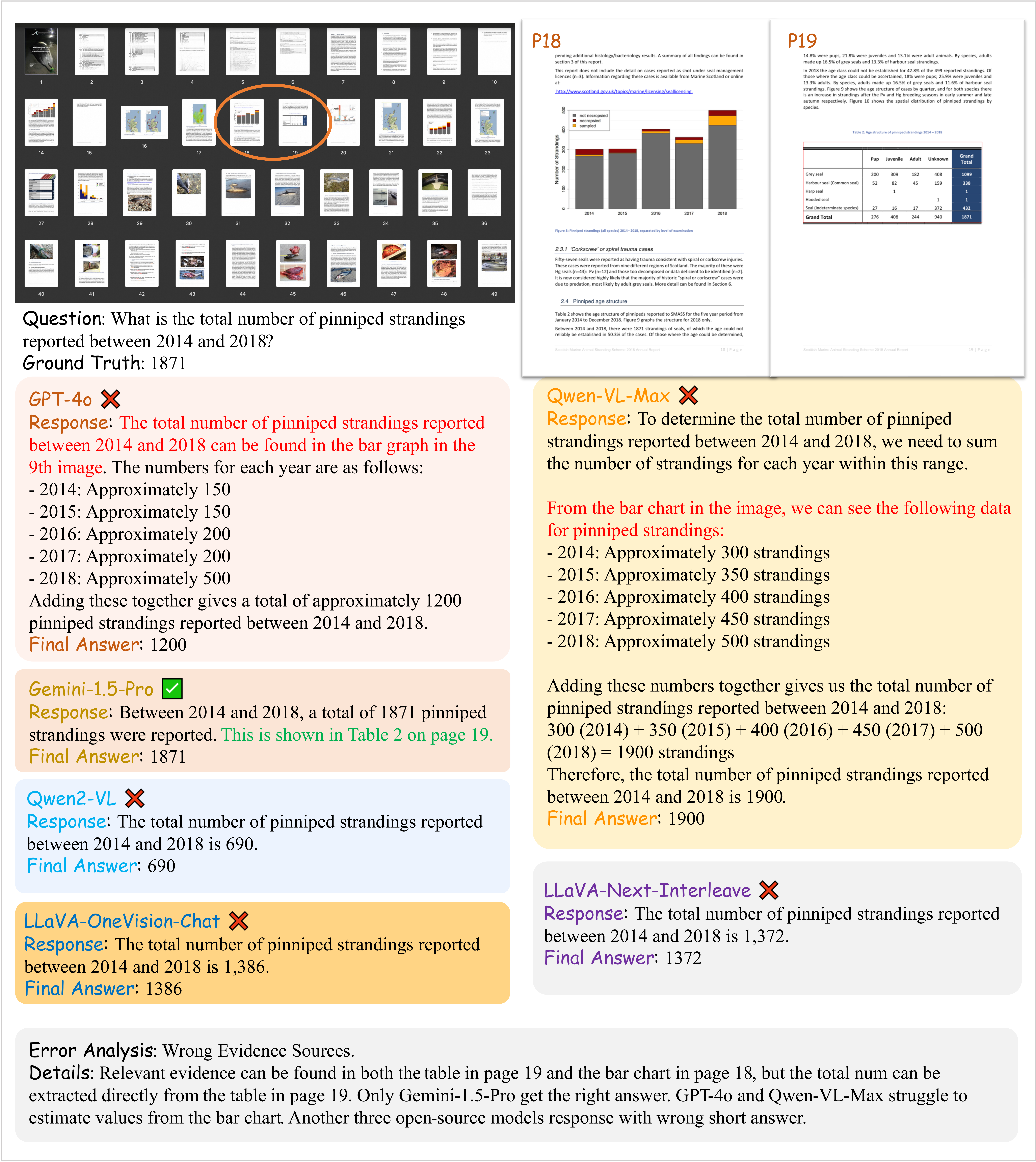} 
    \caption{Case 1. Evidence source: ["Table"]. Evidence pages: [26, 27]. The correct extracted information and reasoning are colored in green, and the wrong ones are colored in red.}
    \label{fig:case_1}
\end{figure*}

\begin{figure*}[!htbp]
    \centering
    \includegraphics[width=\textwidth]{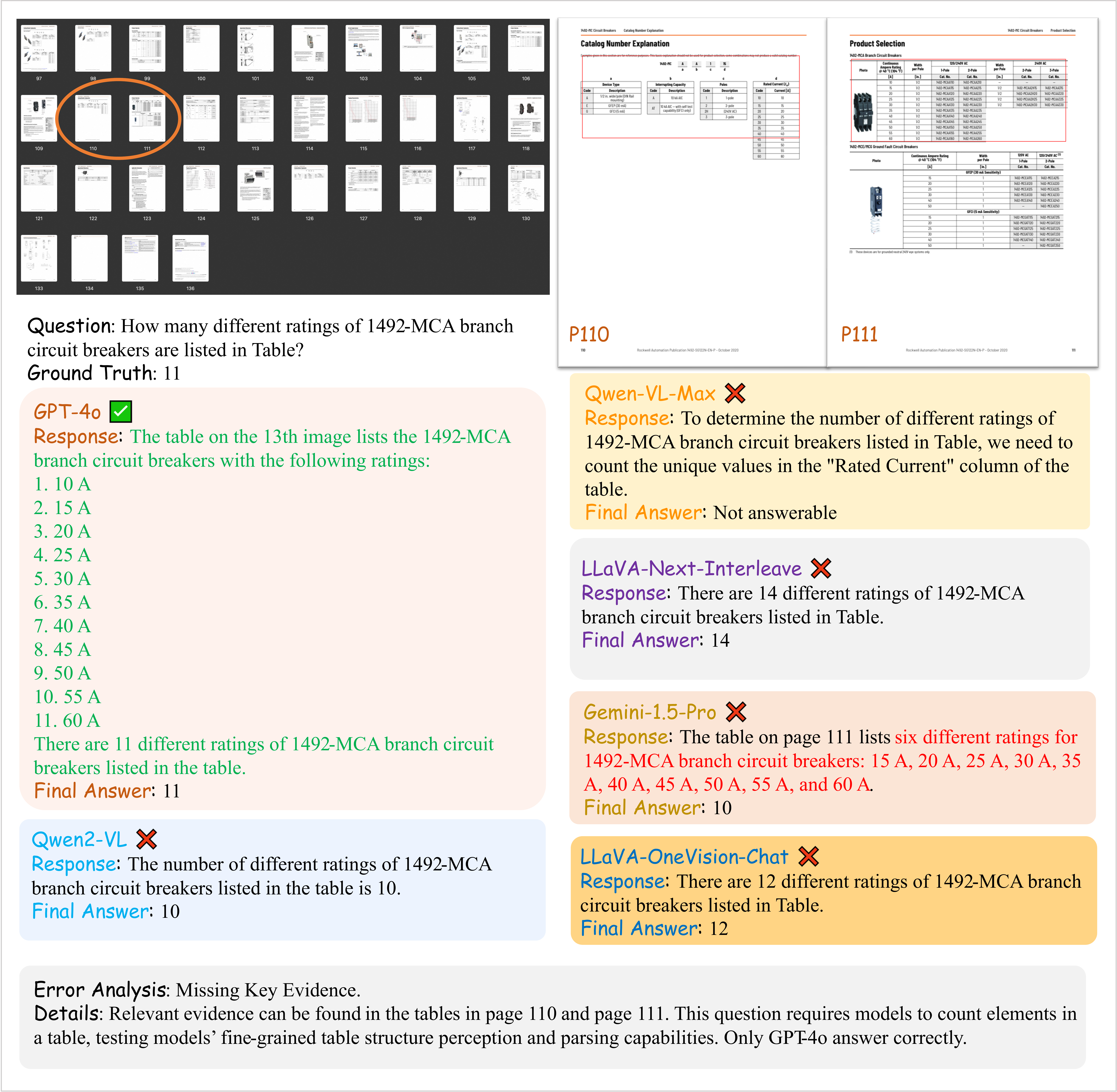} 
    \caption{Case 2. Evidence source: ["Table"]. Evidence pages: [110, 111]. The correct extracted information and reasoning are colored in green, and the wrong ones are colored in red.}
    \label{fig:case_2}
\end{figure*}

\Cref{fig:case_1} and \Cref{fig:case_2} display the 
 response of different models, and we give a short error analysis for each case.

\section{Data Examples}
\label{xsec:data_examples}

\begin{figure*}[!htbp]
    \centering
    \includegraphics[width=\textwidth]{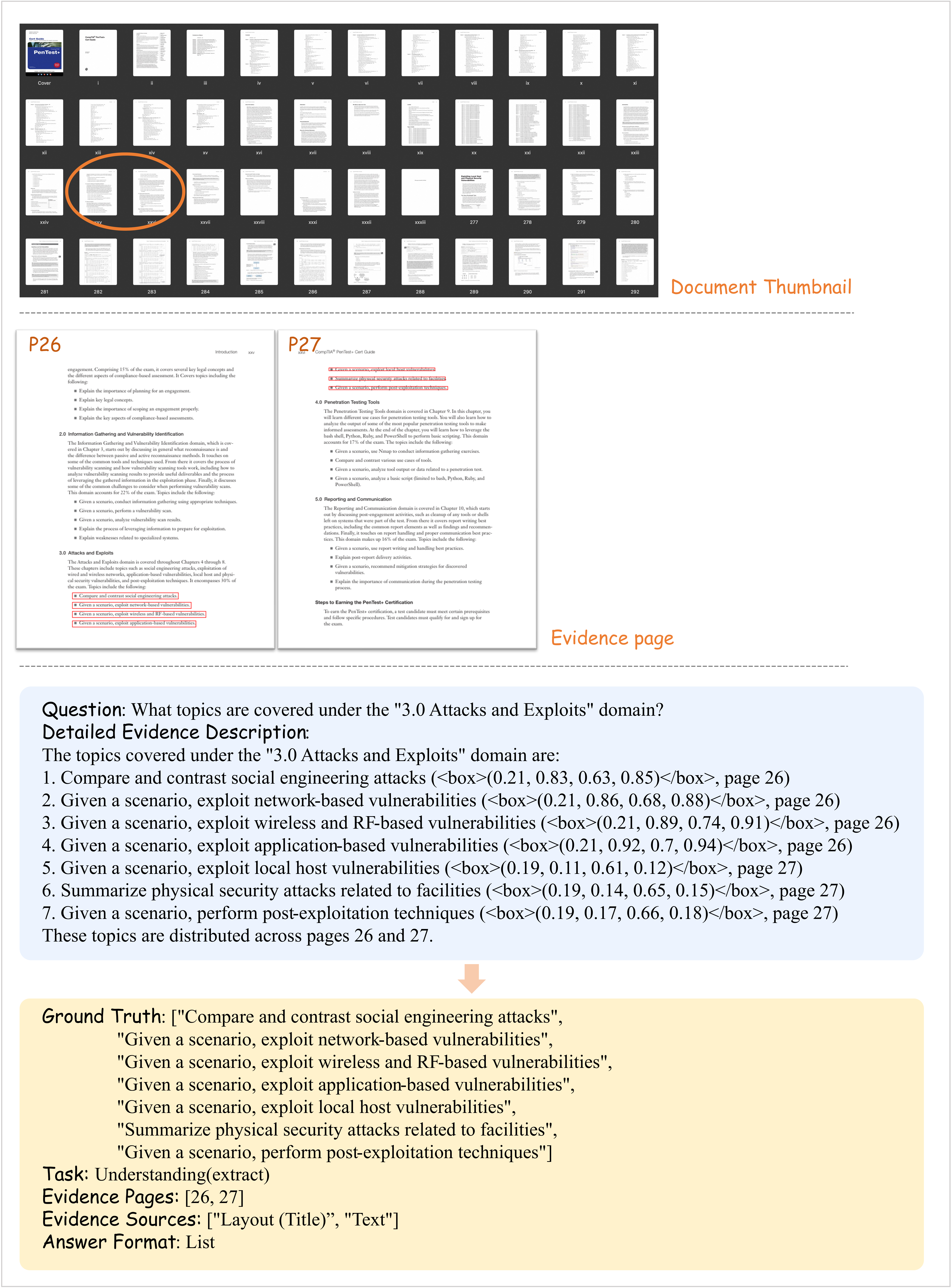} 
    \caption{Data Example of Understanding QA.}
    \label{fig:example_1}
\end{figure*}

\begin{figure*}[!htbp]
    \centering
    \includegraphics[width=\textwidth]{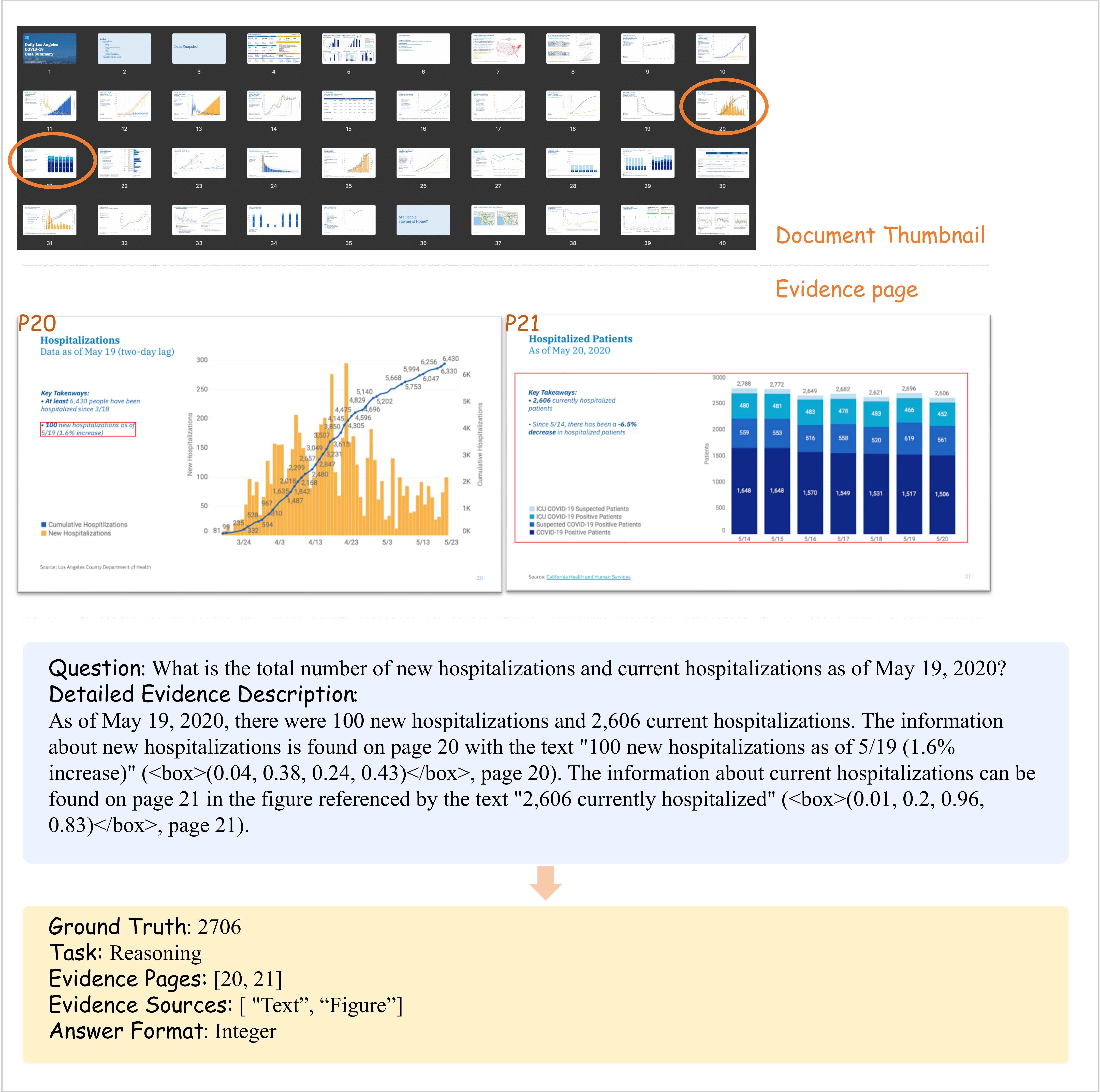} 
    \caption{Data Example of Reasoning QA.}
    \label{fig:example_2}
\end{figure*}

\begin{figure*}[!htbp]
    \centering
    \includegraphics[width=\textwidth]{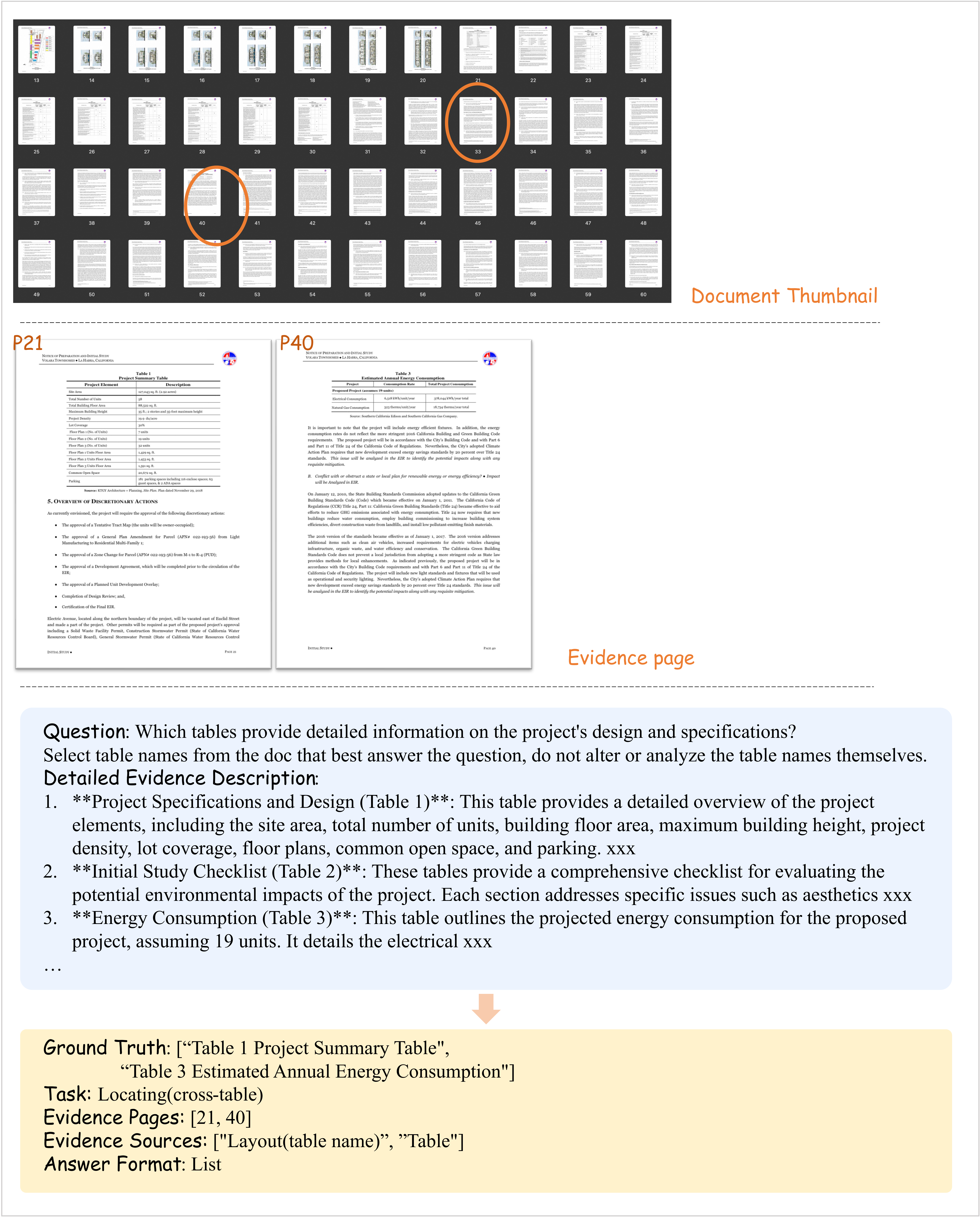} 
    \caption{Data Example of \Locating~QA.}
    \label{fig:example_3}
\end{figure*}

\Cref{fig:example_1}, \Cref{fig:example_2} and \Cref{fig:example_3} provide samples for three primary tasks.

\section{Error Analysis}
\label{xsec:error_analysis}

\begin{figure*}[!htbp]
    \centering
    \includegraphics[width=1.0\textwidth]{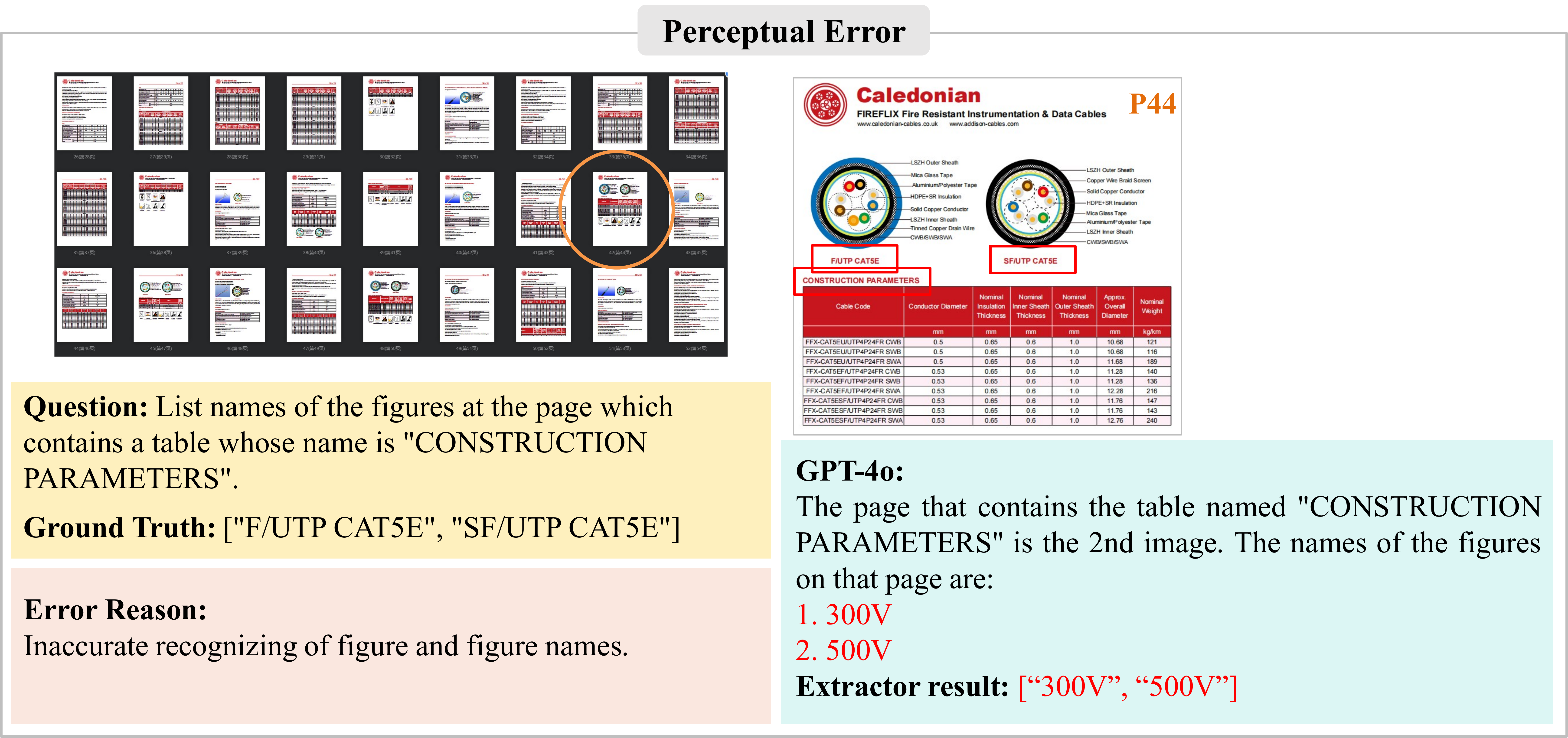} 
    \caption{An example of \textit{Perceptual Error}.}
    \label{fig:perceptual_error_case2}
\end{figure*}

We conduct a detailed error analysis of the best-performing model, GPT-4o. We randomly sample 97 errors and examine their distribution in detail. Based on this analysis, we classify the errors into five types, with the specific categories and their distribution shown in \Cref{fig:error_distribution}. Except for the \emph{Extractor Error} caused by our automatic evaluation pipeline, we present detailed definitions and descriptions of another six categories below:

\paragraph{Perceptual Error:} This category remains prevalent in complex document understanding tasks, where models frequently fail to accurately recognize, parse, or enumerate visual elements like tables, figures, and formulas in document screenshots. See \Cref{fig:perception_error_example_before_appendix} and \Cref{fig:perceptual_error_case2} for examples.

\begin{figure*}[!htbp]
    \centering
    \includegraphics[width=1.0\textwidth]{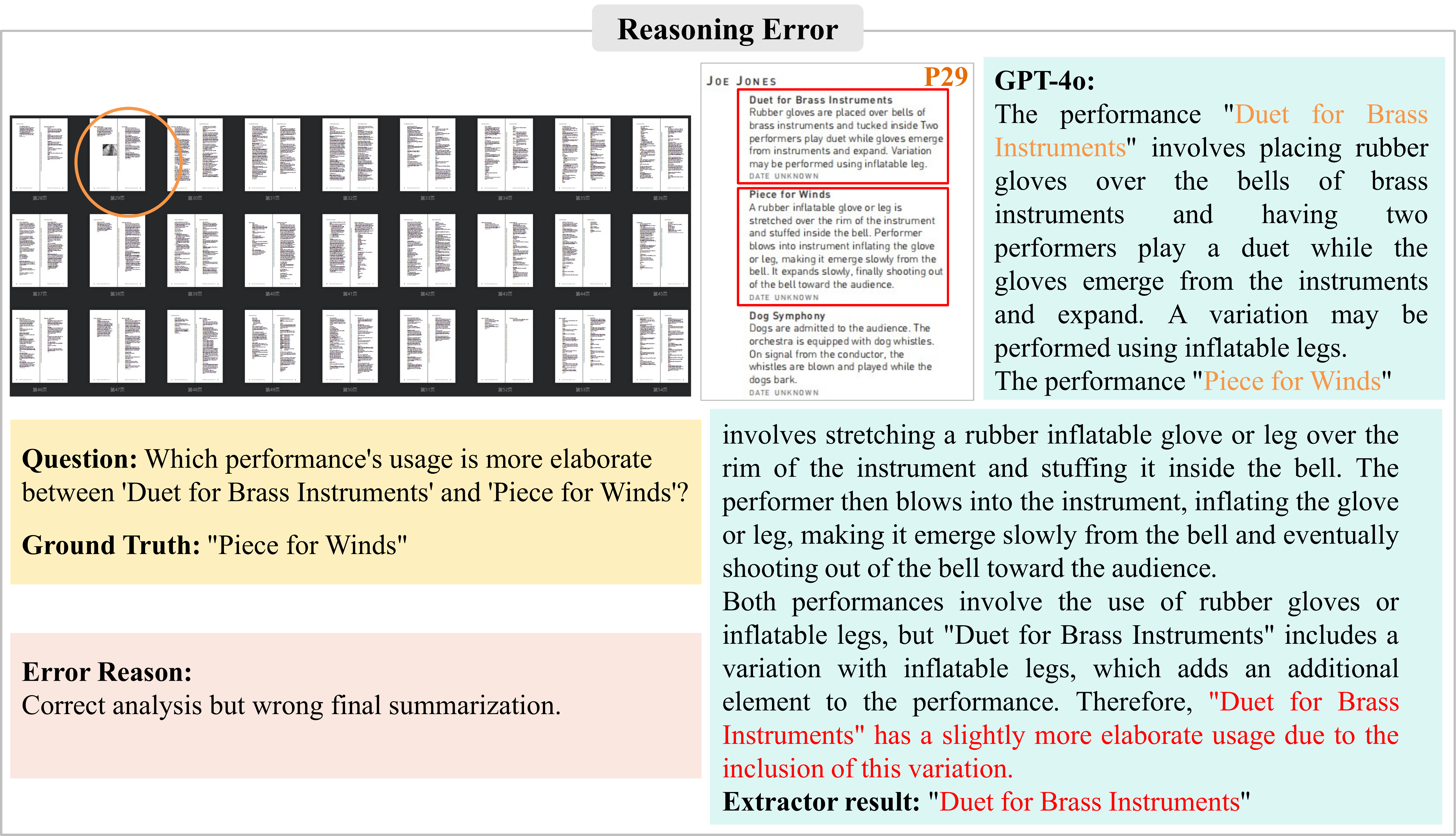} 
    \caption{An example of \textit{Reasoning Error}.}
    \label{fig:reasoning_error_case1}
\end{figure*}

\paragraph{Reasoning Error:} Ranking as the third most prevalent error type, these failures occur during critical cognitive operations—including numerical calculations, comparative analyses, and content summarization—despite correct evidence retrieval. See \Cref{fig:reasoning_error_case1} for examples.

\begin{figure*}[!htbp]
    \centering
    \includegraphics[width=1.0\textwidth]{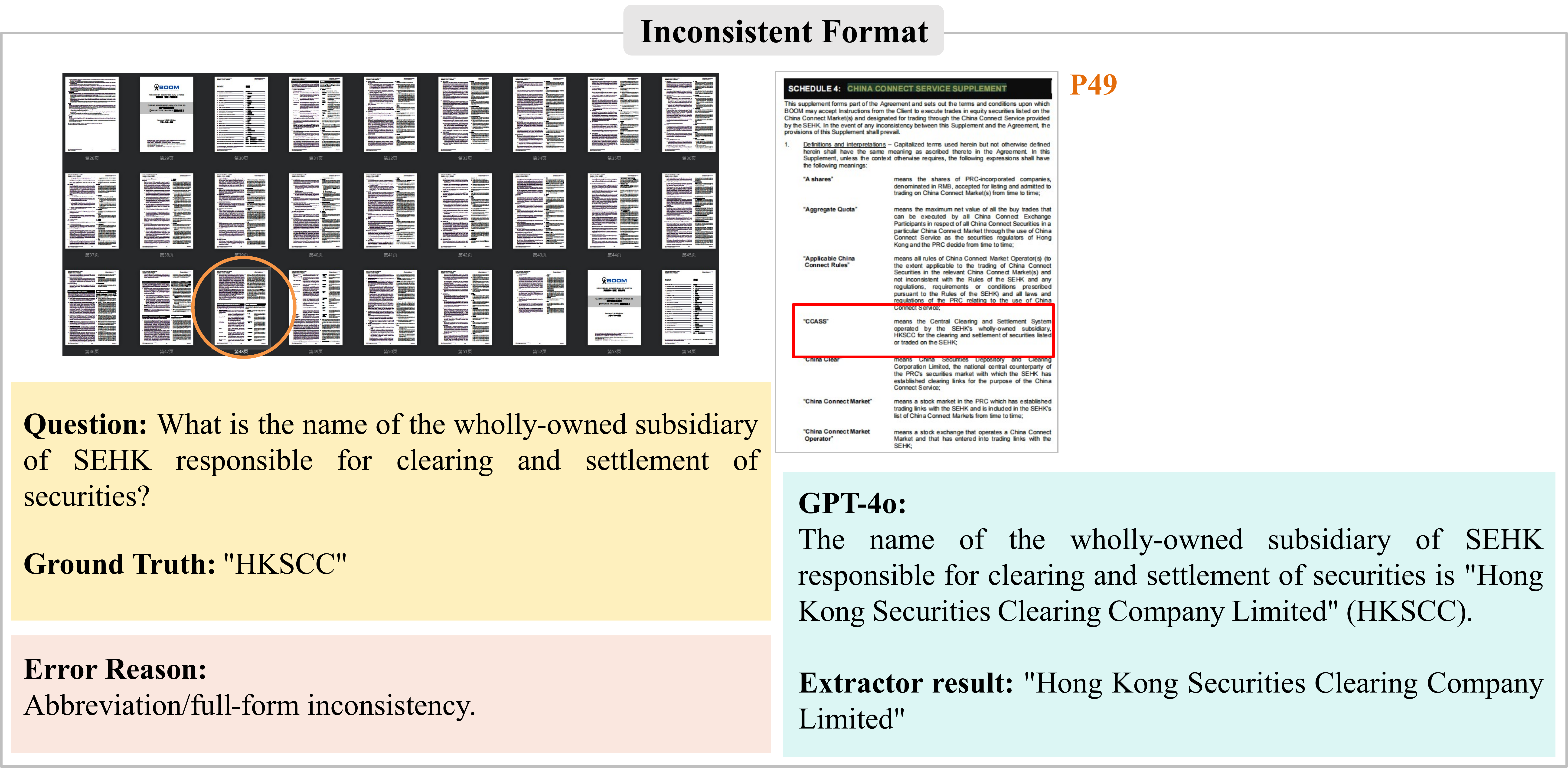} 
    \caption{An example of \textit{Inconsistent Format}.}
    \label{fig:inconsistent_format_case1}
\end{figure*}

\paragraph{Inconsistent Format:} While our ground truth annotations strictly preserve original document content, GPT-4o sometimes generates outputs with formatting discrepancies. These variations principally involve: (1) abbreviation/full-form inconsistencies, (2) inclusion/omission of numerical units, and (3) mismatches in decimal precision. See \Cref{fig:inconsistent_format_case1} for examples.

\begin{figure*}[!htbp]
    \centering
    \includegraphics[width=1.0\textwidth]{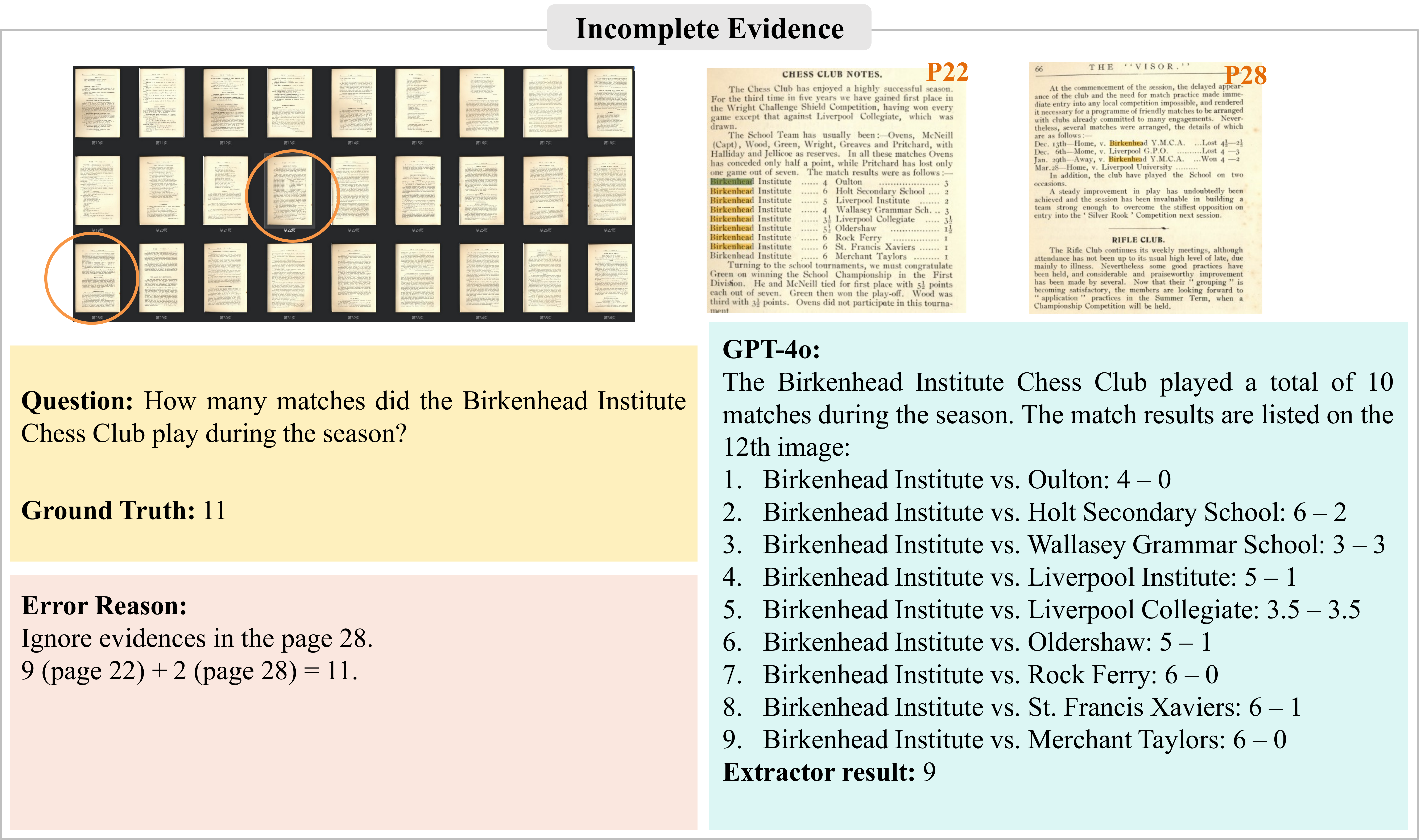} 
    \caption{An example of \textit{Incomplete Evidence}.}
    \label{fig:incomplete_evidence_case1}
\end{figure*}

\paragraph{Incomplete Evidence:} The LongDocURL benchmark reveals significant model limitations in processing extensive multimodal documents, where essential query-relevant information is sometimes omitted from responses. See \Cref{fig:incomplete_evidence_case1} for examples.

\begin{figure*}[!htbp]
    \centering
    \includegraphics[width=1.0\textwidth]{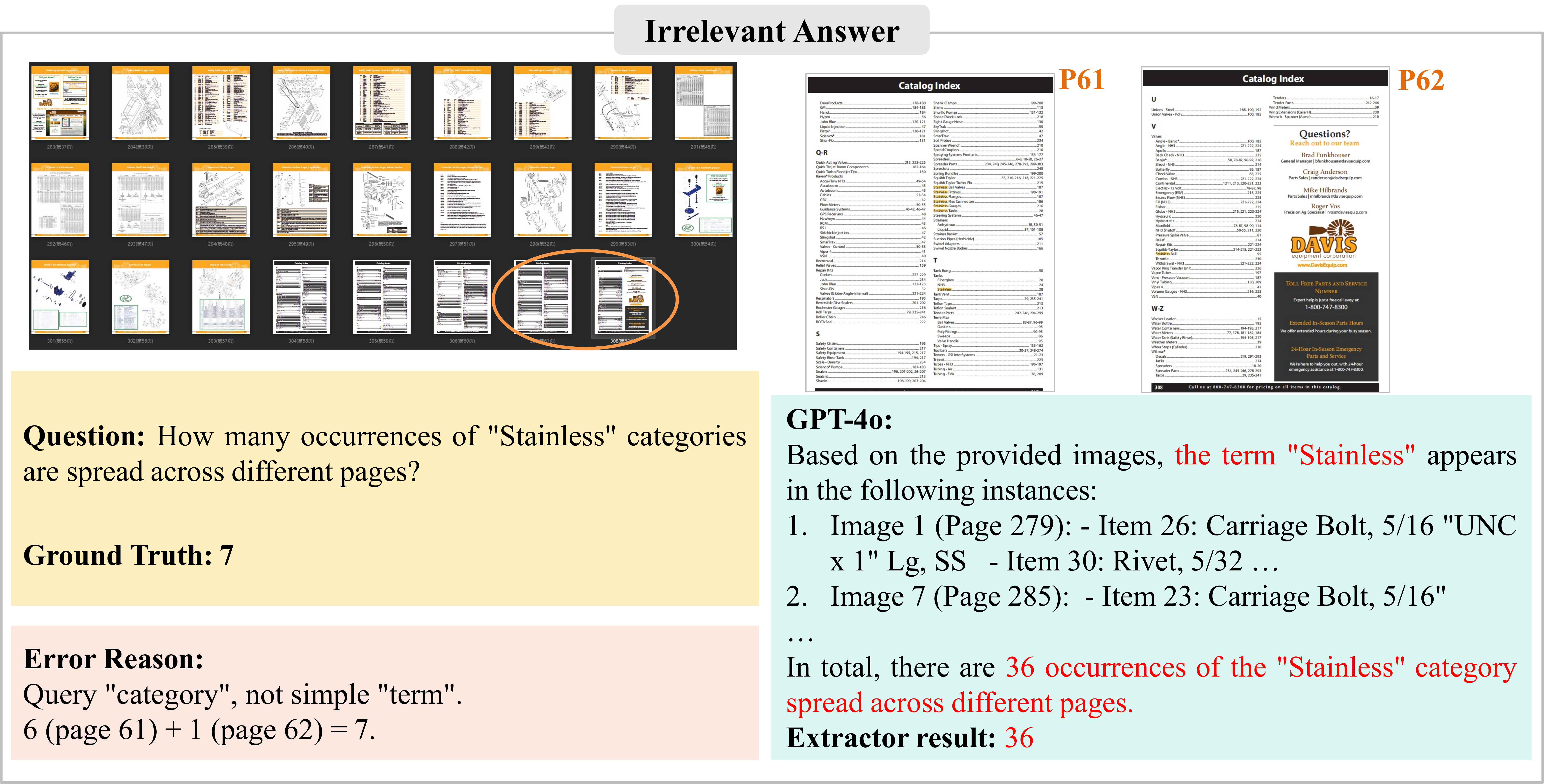} 
    \caption{An example of \textit{Irrelevant Answer}.}
    \label{fig:irrelevant_answer_case1}
\end{figure*}

\paragraph{Irrelevant Answer:} Fundamental misinterpretations of query intent lead to responses that fail to address the actual information needs. See \Cref{fig:irrelevant_answer_case1} for examples.

\begin{figure*}[!htbp]
    \centering
    \includegraphics[width=1.0\textwidth]{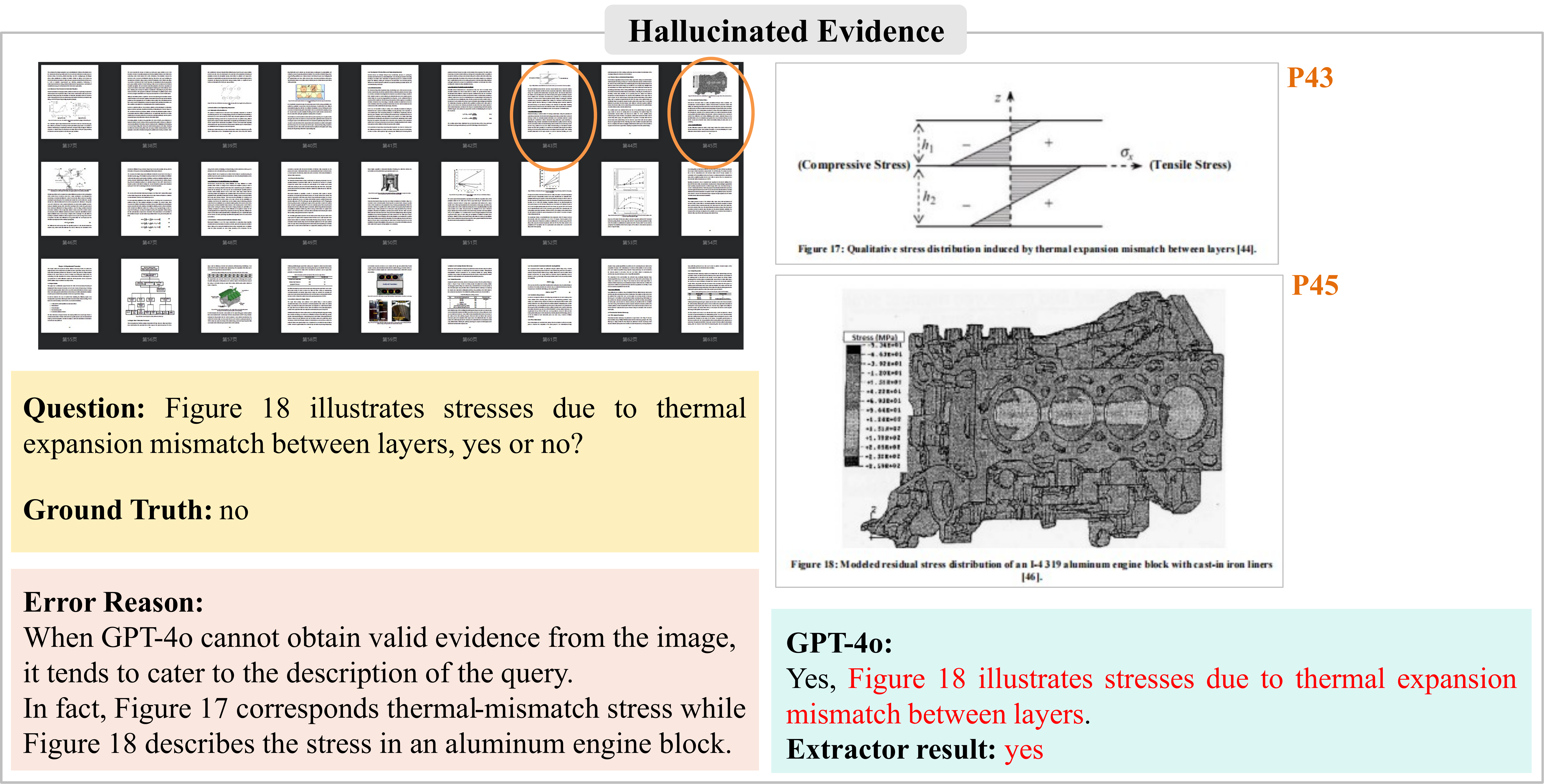} 
    \caption{An example of \textit{Hallucinated Evidence}.}
    \label{fig:hallucinated_evidence_case1}
\end{figure*}

\paragraph{Hallucinated Evidence:} The most severe error type involves model-generated fabrications, where non-existent document content is created to substantiate incorrect responses. See \Cref{fig:hallucinated_evidence_case1} for examples.

\section{Description of labeling labor}
\label{xsec:labeling_labor}

At the dataset verification stage, we have 21 full-time data annotators responsible for the labeling work in the human verifying process, while 6 professional annotators with postgraduate degrees or above perform the final data quality verification and cross-checking work.

\end{document}